\PassOptionsToPackage{table}{xcolor}

\documentclass[10pt,twocolumn,letterpaper]{article}

\usepackage[margin=0.75in]{geometry}
\usepackage[T1]{fontenc}
\usepackage[utf8]{inputenc}
\usepackage{newtxtext}
\usepackage{helvet}
\usepackage{courier}
\usepackage[hyphens]{url}
\usepackage{graphicx}
\usepackage[round]{natbib}
\usepackage{caption}
\usepackage{abstract}
\usepackage{booktabs}
\usepackage{amsmath}
\usepackage{amssymb}

\usepackage{newtxmath}
\usepackage{multirow}
\usepackage{tikz}
\usepackage{pgfplots}
\pgfplotsset{compat=1.18}
\usetikzlibrary{patterns,calc,positioning,arrows.meta,shapes.geometric,fit,backgrounds}

\usepackage[colorlinks=true,allcolors=blue,breaklinks=true]{hyperref}

\definecolor{navy}{HTML}{1E2761}
\definecolor{coral}{HTML}{E85D3B}
\definecolor{teal}{HTML}{2C7A7B}
\definecolor{gold}{HTML}{D69E2E}
\definecolor{slate}{HTML}{4A5568}
\definecolor{lightnavy}{HTML}{E6EAF5}
\definecolor{lightcoral}{HTML}{F8D6C5}
\definecolor{rowgray}{HTML}{ECF0F5}

\title{\vspace{-1.5em}\bfseries CoTinyVLA: Chain-of-Thought Distillation for a\\ Sub-Billion-Parameter Vision-Language-Action Model\vspace{-0.5em}}

\author{%
Minhyeok Lee$^{*}$\quad
Chiyoung Kim\quad
Chanhoe Gu\quad
Seongrok Kim\\
Sanghyuk Roy Choi\quad
Donghwan Hwang\quad
Donghun Ryu\quad
Seokhyun Kim\\[0.4em]
\normalsize Chung-Ang University, Republic of Korea\\
\normalsize \texttt{mlee@cau.ac.kr}
}
\date{}

\begin{document}

\twocolumn[\begin{@twocolumnfalse}
\maketitle
\begin{onecolabstract}
Vision-Language-Action (VLA) models translate natural-language commands into robot action sequences, but leading systems on the LIBERO-Plus robustness benchmark use three- to seven-billion-parameter backbones whose memory demands can exceed embedded robotic budgets. We present CoTinyVLA, a 0.9B-parameter action model on a Qwen3.5-0.8B backbone that obtains that robustness by structuring supervision instead of enlarging the model. Three components target different axes of the problem: dual-view temporal input of 16 history frames per step with textual camera and time markers; hierarchical chain-of-thought (CoT) distillation from a 35B teacher into an episode-level \textsc{Plan} and a chunk-level \textsc{Think} span over task phase, gripper state and next subaction; and paraphrase augmentation expanding 40 base commands into 800 variants. On LIBERO-Plus, spanning 10{,}030 perturbed tasks across seven perturbation dimensions, CoTinyVLA reaches 90.8\% on Spatial, 87.3\% on Object, 86.6\% on Goal and 80.7\% on Long, leading the strongest 7B baseline on all four suites by 4.7, 2.8, 15.9 and 3.0 points, with every margin interval excluding zero. The gains concentrate on the hardest axes of the benchmark: across the eleven published baselines none exceeds 53.2\% on Robot Initial States in any suite, whereas CoTinyVLA reaches 73.6\% on Goal against 39.9\% for the strongest baseline. Ablations show the three components to be separable by perturbation axis, and at a matched image budget how frames are divided between the two cameras and across time accounts for 8.6 points on its own. Closed-loop inference peaks at 2.25\,GiB of allocated GPU memory, and paired interventions show the episode \textsc{Plan} to be load-bearing: replacing it with an empty or contradictory span costs 40 to 45 points of success. Structured supervision across temporal, reasoning and linguistic axes thus lets a 0.9B backbone exceed all of them on all four suites. Code: \url{https://github.com/BrainJellyPie/CoTinyVLA}.
\end{onecolabstract}
\vspace{1em}
\end{@twocolumnfalse}]
{\renewcommand{\thefootnote}{$*$}\footnotetext{Corresponding author: \texttt{mlee@cau.ac.kr}}}

\section{Introduction}
\label{sec:intro}

Vision-language-action models have emerged as a unified interface between natural-language instructions and embodied control. Recent open systems such as OpenVLA \citep{kim2025openvla}, OpenVLA-OFT \citep{kim2025openvlaoft}, $\pi_0$ \citep{black2024pi0}, NORA \citep{hung2025nora}, UniVLA \citep{bu2025univla}, WorldVLA \citep{cen2025worldvla}, and RIPT-VLA \citep{tan2025riptvla} achieve high success rates on simulated manipulation suites, and until now the best results on the LIBERO-Plus robustness benchmark \citep{fei2025liberoplus} came from this group. Most reuse a pretrained vision-language backbone of three to seven billion parameters.

This parameter range is difficult to accommodate on the hardware where language-conditioned robots are deployed: mobile manipulators, humanoids, and assistive platforms rely on embedded accelerators with strict memory budgets, whereas a 7B-scale VLA in \texttt{bfloat16} typically requires 20 gigabytes or more of GPU memory. On LIBERO-Plus their advantage lies in robustness to visual, kinematic and linguistic perturbation, which raises two questions: whether targeted training signal can give a compact model the same benefit, and which axis of the robustness problem each form of supervision addresses. We build a sub-billion VLA around three components, each aimed at a different axis:

\textbf{Dual-view temporal input.} At each control step the model receives eight third-person and eight wrist frames, each prefixed by an explicit textual frame index. This gives the compact backbone access to motion across time and cameras, a cue that complements the pretrained representation.

\textbf{Hierarchical chain-of-thought (CoT) distillation.} A 35B vision-language teacher produces two levels of structured reasoning: an episode-level \textsc{Plan} decomposing the instruction into ordered subgoals, and a chunk-level \textsc{Think} stating three discrete attributes (manipulation phase, gripper state, next subaction). The student predicts both the reasoning tokens and the action chunk, so the reasoning span acts as an auxiliary structured prediction objective that adds no inference-time parameters.

\textbf{Paraphrase-based instruction augmentation.} We expand 40 base instructions into 800 paraphrases (verb swaps, synonyms, politeness variation), and each sample's instruction is replaced with a paraphrase with probability~0.8, improving robustness to lexical variation in the instruction.

We build these ideas into \textbf{CoTinyVLA}, a 0.9B-parameter language-grounded action model on a Qwen3.5-0.8B backbone \citep{qwen2026qwen35}, fine-tuned on the four LIBERO-Plus suites for two epochs and evaluated on both LIBERO-Plus and the original LIBERO benchmark \citep{liu2023libero}.

CoTinyVLA reaches 90.8\%, 87.3\%, 86.6\% and 80.7\% on the Spatial, Object, Goal and Long suites, the highest value on each. Against the strongest 7B baseline the margins are 4.7, 2.8, 15.9 and 3.0 points and every interval for the difference excludes zero, so a 0.9B model leads systems eight times its size on all four suites. The margins are largest where the baselines are weakest. On Goal, the two axes on which the strongest baseline scores lowest are Robot Initial States (39.9\%) and Language Instructions (56.6\%); CoTinyVLA reaches 73.6\% and 83.7\% on them.

The three components are also separable, not just additive. Removing paraphrase augmentation moves the Language Instructions axis by 12.3 points and leaves the six non-linguistic axes unchanged; removing chain-of-thought distillation costs most on the physical-state axes; removing the temporal history costs most under kinematic perturbation. At a fixed image budget the allocation of frames across cameras and time is itself worth 8.6 points, so input structure is a design variable in its own right, not a fixed cost of the architecture.

\textbf{Contributions.} (1)~CoTinyVLA, a 0.9B-parameter VLA that exceeds the strongest 7B baselines on all four LIBERO-Plus suites, by 4.7 to 15.9 points, within a measured 2.25\,GiB inference budget. (2)~A two-level reasoning schema in which the episode \textsc{Plan} carries task intent, is shown by test-time intervention to condition the emitted actions, and can be cached across the episode, halving inference cost. (3)~A decomposition of the robustness problem by perturbation axis: controlled ablations attribute each component to a distinct group of perturbations, test-time interventions separate the two levels of the reasoning hierarchy, and input structure emerges as a design variable in its own right, worth 8.6 points at a fixed image budget.

\section{Related Work}
\label{sec:related}

\paragraph{Vision-language-action models.}
VLAs cast robot control as conditional sequence prediction grounded in a pretrained vision-language backbone. Early systems such as RT-1 \citep{brohan2023rt1} established the paradigm at billion-parameter scale; the current LIBERO-Plus frontier \citep{fei2025liberoplus} is dominated by 3--7B systems including OpenVLA \citep{kim2025openvla}, its efficient fine-tuning recipe OpenVLA-OFT \citep{kim2025openvlaoft}, $\pi_0$ \citep{black2024pi0} with action tokenisation \citep{pertsch2025fast}, UniVLA \citep{bu2025univla}, WorldVLA \citep{cen2025worldvla} and RIPT-VLA \citep{tan2025riptvla}. A complementary line targets reduced footprints: TinyVLA \citep{wen2024tinyvla} and SmolVLA \citep{shukor2025smolvla} build data-efficient compact backbones, FLOWER \citep{reuss2025flower} reaches strong LIBERO performance at 0.9B, and NORA \citep{hung2025nora} is an open compact generalist; Table~\ref{tab:standard} lists fourteen baselines below one billion parameters. That a compact VLA can be competitive is therefore the premise of this work, not its finding. The question we take up is what has to be supplied in place of scale, and the answer we report is that supplying it is enough to move a 0.9B model from competitive to ahead of every baseline in that comparison on all four LIBERO-Plus suites.

Closest to our input design, CronusVLA \citep{li2025cronusvla} conditions a compact VLA on a multi-frame visual history. CoTinyVLA keeps the backbone scale similarly small but supervises the resulting representation with hierarchical CoT and pairs a third-person frame history with a wrist-camera frame history. Compute-adaptive designs such as looped or early-exit transformers approach the same footprint problem from the other side, reducing the compute spent per parameter at fixed supervision; holding parameters and depth fixed and changing what the supervision contains is orthogonal to that, so the two compose.

\paragraph{Chain-of-thought reasoning and distillation.}
Chain-of-thought prompting \citep{wei2022chain} has been adapted to robot control. ECoT \citep{zawalski2024ecot} emits a free-form natural-language reasoning trace before each low-level action, at a scale where the backbone can produce such a trace unaided. The NLP literature has independently shown that CoT rationales from a large teacher can be distilled into a substantially smaller student \citep{hsieh2023distilling,ho2023reasoningteachers}. We follow that route: a 35B vision-language teacher supplies hierarchical \textsc{Plan} and \textsc{Think} traces during training, and the 0.9B student, which learns to predict both the reasoning tokens and the demonstrated action chunk, generates them on its own at inference.

\paragraph{Benchmarks and instruction augmentation.}
We evaluate on LIBERO \citep{liu2023libero}, a four-suite lifelong-learning manipulation benchmark, and on its recent robustness extension LIBERO-Plus \citep{fei2025liberoplus}, which perturbs each task along seven dimensions including a Language Instructions axis. To improve robustness along this axis, we draw on text-augmentation methods from NLP such as surface-form perturbation \citep{wei2019eda} and large-scale paraphrase generation \citep{hu2019parabank}.

\section{Method}
\label{sec:method}

\begin{figure*}[!t]
\centering
\includegraphics[width=\textwidth]{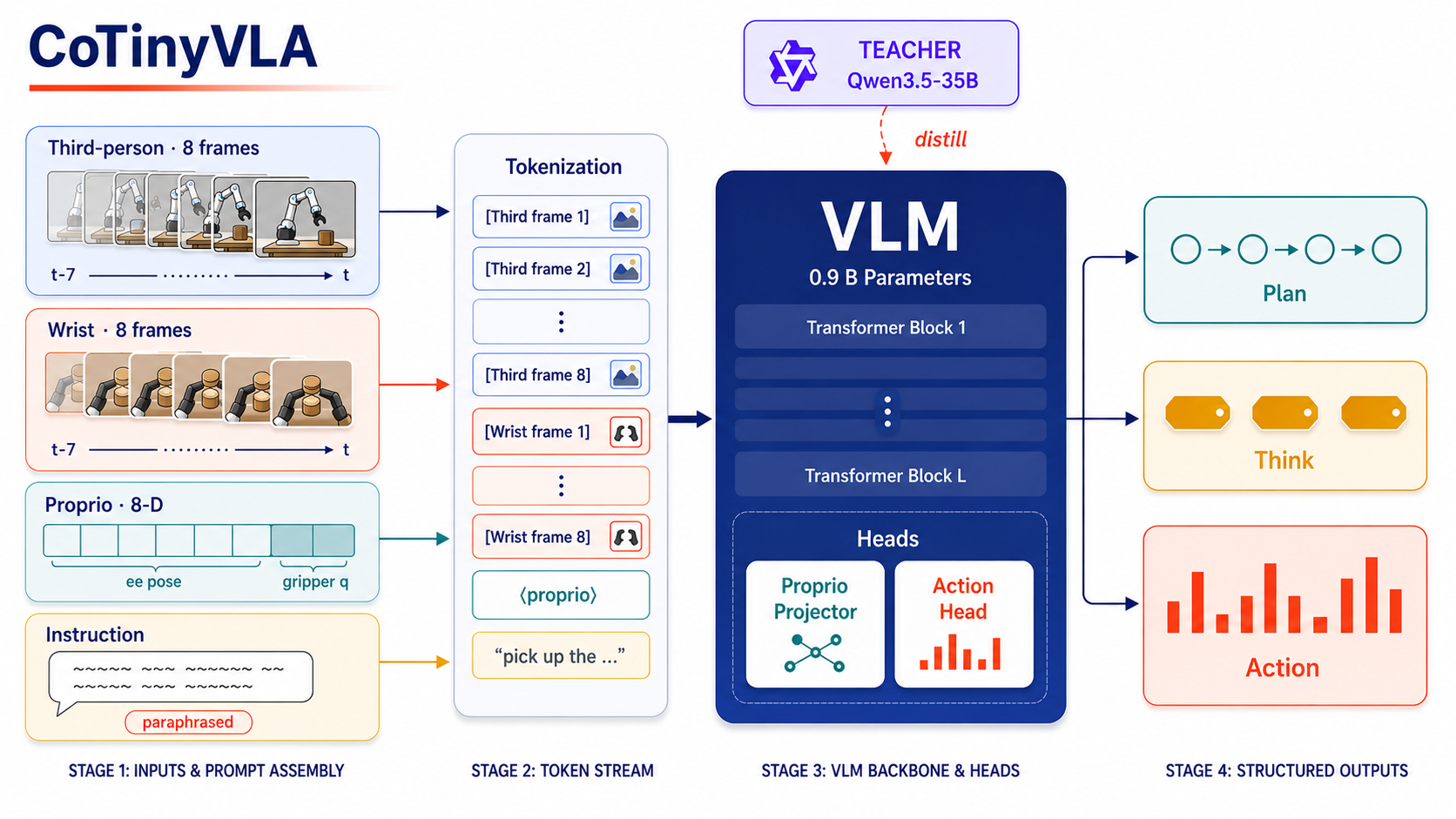}
\caption{Overview of CoTinyVLA. The model is organised as a four-stage pipeline. Stage~1 assembles four input modalities (a third-person frame history, a wrist frame history, an 8-dimensional proprioception vector, and a paraphrased natural-language instruction). Stage~2 tokenises every input into a single linear token stream with explicit per-view per-step textual markers. Stage~3 is the Qwen3.5-0.8B vision-language backbone (approximately 0.9B parameters in total) together with a proprioception projector and an action head. Stage~4 produces three structured outputs: an episode-level \textsc{Plan}, a chunk-level \textsc{Think} with structured Phase, Gripper and Next tags, and an 8-step action chunk. The \textsc{Plan} and \textsc{Think} reasoning tokens are supervised by a 35B vision-language teacher (top, dashed coral distillation arrow); actions are supervised directly by demonstration.}
\label{fig:arch}
\end{figure*}

\subsection{Architecture}
\label{sec:method-arch}

CoTinyVLA is a single vision-language-action transformer with a proprioception projector on the input side and an action head on the output side. The backbone is Qwen3.5-0.8B; counting the additional embeddings introduced for our control tokens, the proprioception projector, and the action head, the final model has approximately 0.9B parameters. At each control step the model receives:

\begin{itemize}
\itemsep0pt
\item Eight third-person frames $\{I^{\text{3p}}_{t-7}, \ldots, I^{\text{3p}}_{t}\}$;
\item Eight wrist frames $\{I^{\text{wr}}_{t-7}, \ldots, I^{\text{wr}}_{t}\}$;
\item An 8-dimensional proprioception vector (end-effector pose plus two gripper joint positions);
\item A natural-language instruction.
\end{itemize}

Images are encoded by the backbone vision encoder at $224{\times}224$ resolution. Each image is preceded by an explicit text marker, ``\texttt{[Third frame $i$]}'' or ``\texttt{[Wrist frame $i$]}'' with $i\in\{1,\ldots,8\}$, so that temporal and camera-view indices are available as explicit token context, not only implicitly through the vision stream. Proprioception is projected to the backbone hidden size by a small two-layer MLP with $\sim$1\,M parameters and prepended to the language stream as a single soft token. The action head consumes the final hidden state at a fixed query position and produces an 8-step action chunk $\mathbf{a}_{t:t+7}\in\mathbb{R}^{8\times 7}$ (six end-effector deltas plus one gripper command per step). Figure~\ref{fig:arch} gives an overview.

An eight-frame window per view spans roughly 0.4\,s of history at the typical LIBERO control frequency of 20\,Hz, which covers a complete grasp onset while keeping the resulting context within the token budget at $224{\times}224$ resolution for the compact backbone. Eight frames balance two factors: a single frame is insufficient for kinematic inference, while longer windows increase token count. The two viewpoints are complementary, not redundant: the third-person view provides global scene layout while the wrist view provides a close-up of gripper--object contact. Feeding both lets the small backbone attend to whichever view is more informative for the current step; at a matched image budget, splitting it across the two views outperforms spending it on either camera alone. The explicit textual frame markers are an inductive bias: they encode image order directly in the token stream, so no additional temporal positional embeddings are introduced. The visual history is inexpensive here: each additional image adds roughly 32\,MiB of activation memory and 6.5\,ms of forward time, so profiling the forward pass alone, the full 16-image input raises peak allocated memory from 1.76\,GiB to 2.20\,GiB, a small increment relative to the weights.

\subsection{Hierarchical Chain-of-Thought}
\label{sec:method-cot}

We introduce four new tokens into the backbone vocabulary: \texttt{<plan>}, \texttt{</plan>}, \texttt{<think>}, \texttt{</think>}. The backbone autoregressively generates two reasoning structures before the action chunk.

\textsc{Plan} is produced once per episode, immediately after the instruction is consumed. It breaks down the instruction into an ordered list of subgoals in natural language.

\textsc{Think} is produced once per action chunk and is constrained to a fixed three-slot schema:
\begin{quote}\footnotesize
\texttt{Phase: <step k: short description>}\\
\texttt{Gripper: <OPEN|CLOSED|PARTIALLY\_CLOSED>}\\
\texttt{Next: <short action description>}
\end{quote}
\textsc{Phase} indexes the current subaction against the episode-level \textsc{Plan}, so its vocabulary is bounded by the \textsc{Plan} and not fixed in advance, while \textsc{Gripper} is drawn from a closed vocabulary and can therefore be validated automatically against proprioception during label generation. The three slots act as an auxiliary structured prediction target alongside the action loss, and the action head is conditioned on the representation that follows the span. We constrain \textsc{Think} to these slots instead of allowing free-form reasoning: the fixed-slot schema bounds the length of the span and concentrates supervision on three variables tied directly to action correctness.

The \textsc{Plan}-then-\textsc{Think} split matches the two timescales of a manipulation task. \textsc{Plan} is a stable representation of \emph{intent} that stays fixed as the robot moves, so generating it once per episode gives every chunk in the episode a single shared representation of the task; \textsc{Think} is a representation of \emph{state} that must update as the action progresses, so it is produced at every chunk. The split also pays at inference: the \textsc{Plan} can be emitted once and cached, which halves steady-state inference cost, whereas a flat per-chunk trace has no equivalent saving. Test-time interventions confirm that the episode-level component is load-bearing: replacing the \textsc{Plan} with an empty or contradictory one costs 40 to 45 points of success (supplementary material).

\subsection{Teacher and Distillation}
\label{sec:method-distill}

\textsc{Plan} and \textsc{Think} labels are produced by a 35B Mixture-of-Experts vision-language teacher (Qwen3.5-35B-A3B) run with 4-bit quantisation during label generation. For each demonstration, the teacher generates one \textsc{Plan} from the instruction and the first frame, and 10 \textsc{Think} labels at uniformly spaced chunks along the episode; chunks in between reuse the most recent label, so every chunk carries a label at ten teacher queries per demonstration.

Two aspects of the labelling matter. The teacher prompt for \textsc{Think} reads the proprioception vector at the corresponding chunk and provides a derived gripper hint (\textsc{OPEN}, \textsc{PARTIALLY\_CLOSED}, or \textsc{CLOSED} based on the inter-finger joint gap). This ties the Gripper attribute to physical state instead of image inference, avoiding ambiguity in static frames where opening and closing are visually similar. The action chunk label is taken directly from the demonstration, not from the teacher; the teacher provides only the reasoning trace. The student is therefore learning \emph{reasoning} from a large vision-language model while learning \emph{actions} from demonstration trajectories.

The training objective is a weighted sum of an action regression loss and a language-model cross entropy over reasoning tokens:
\begin{equation}
\mathcal{L} = \alpha \, \mathcal{L}_{\text{act}}(\hat{\mathbf{a}}, \mathbf{a})
            + \beta \, \mathcal{L}_{\text{lm}}(\hat{\mathbf{y}}_{\text{cot}}, \mathbf{y}_{\text{cot}}),
\end{equation}
with $\alpha=1.0$ and $\beta=0.1$, so that the action objective remains dominant. $\mathcal{L}_{\text{act}}$ is an L1 loss over the 8-step, 7-DoF chunk.

\subsection{Paraphrase-Based Instruction Augmentation}
\label{sec:method-aug}

The four LIBERO-Plus training suites jointly use 40 base instructions, a narrow set of surface forms. To prevent the small student from overfitting the surface form of these instructions, we expand each base instruction into 20 paraphrases, producing 800 natural variants. Paraphrases include verb substitution (\emph{pick up} $\to$ \emph{grab, lift, fetch, retrieve}), object synonym substitution (\emph{black bowl} $\to$ \emph{dark bowl, dark-coloured bowl}; \emph{stove} $\to$ \emph{stovetop, cooktop}), and politeness variation (direct, ``\emph{could you}'', ``\emph{would you mind}'', ``\emph{I'd like}''). At training time, each sample's instruction is replaced by a random paraphrase with probability~0.8; the original instruction is retained otherwise. Paraphrases are used only during training; evaluation always uses the unperturbed reference instruction or the perturbed instruction defined by the benchmark. Lexical coverage statistics for the resulting set are reported in the supplementary material.

The augmentation is motivated by the LIBERO-Plus Language Instructions perturbation, which holds the rest of the scene fixed and rewrites only the natural-language command. Without augmentation, a compact student that has only ever seen 40 surface forms can overfit to those forms, and paraphrasing removes that reliance: within our ablation sweep, the Language Instructions axis falls by 12.3 points when the augmentation is removed, while the six non-linguistic axes are unchanged. The 0.8 swap probability is set high so that the original surface forms do not dominate training; retaining the original instruction with probability 0.2 preserves exposure to the reference wording used at evaluation.

\subsection{Training}
\label{sec:method-training}

The student is fine-tuned on the union of the four LIBERO-Plus suites for two epochs starting from the pretrained Qwen3.5-0.8B backbone (approximately 2.4\,M action-chunk samples). Optimisation uses AdamW with a backbone learning rate of $2\times 10^{-5}$ and a head learning rate of $1\times 10^{-4}$, a linear warmup over 3\% of steps, and gradient checkpointing with FlashAttention-2. Training uses seed 42 and evaluation seed 7, with a deterministic per-episode seed derived from the suite and task index. Training is performed on a single node with eight NVIDIA H100 GPUs.

\subsection{Inference}
\label{sec:method-infer}

At inference, the model produces a \textsc{Plan} once per episode and a \textsc{Think} plus 8-step action chunk on every forward pass; the chunk is executed open-loop for eight environment steps before the next forward pass. Reasoning tokens are decoded greedily with a 120-token cap. With the model generating its own reasoning, a steady-state inference step on an L40S takes 1.37\,s at the median, of which 76\% is autoregressive generation, not the visual stream or the backbone forward pass, and peaks at 2.25\,GiB of allocated memory including the generation KV cache. Caching the \textsc{Plan} for the episode reduces a steady-state span from 70 tokens to 26 and halves that cost. The supplementary material reports the full decomposition and a token-budget sweep.

The 8-step action chunk spreads the cost of a \textsc{Plan}+\textsc{Think} forward pass over eight environment steps, reducing the number of reasoning generations per environment step by a factor of eight relative to per-step CoT decoding. We decode the CoT tokens greedily so that the reasoning span, and the action chunk conditioned on it, is deterministic at evaluation time.

\begin{table*}[t]
\centering
\fontsize{9}{10.5}\selectfont
\setlength{\tabcolsep}{3.5pt}
\renewcommand{\arraystretch}{1.05}
\begin{tabular}{l r r r r r r r r r}
\toprule
\textbf{Model} & \textbf{Params} & \textbf{Camera} & \textbf{Robot} & \textbf{Lang.} & \textbf{Light} & \textbf{Backgr.} & \textbf{Noise} & \textbf{Layout} & \textbf{Total} \\
\midrule
OpenVLA \citep{kim2025openvla} & 7B & 0.0 & 3.7 & 27.7 & 12.3 & 50.4 & 12.0 & 40.7 & 19.4 \\
OpenVLA-OFT \citep{kim2025openvlaoft} & 7B & 88.3 & 40.0 & 80.5 & 98.3 & 97.3 & 96.3 & 93.9 & 84.0 \\
OpenVLA-OFT$_w$ \citep{kim2025openvlaoft} & 7B & 8.8 & 39.7 & 83.6 & 88.4 & 99.2 & 55.3 & 82.7 & 62.5 \\
OpenVLA-OFT$_m$ \citep{kim2025openvlaoft} & 7B & 55.3 & 19.7 & 92.7 & \textbf{100.0} & 92.3 & 85.2 & 94.5 & 75.4 \\
NORA \citep{hung2025nora} & 3B & 4.3 & 50.9 & 63.8 & 66.8 & 65.5 & 12.5 & 84.6 & 47.6 \\
WorldVLA \citep{cen2025worldvla} & 7B & 0.0 & 44.3 & 46.3 & 65.1 & 19.8 & 11.7 & 46.1 & 32.5 \\
UniVLA \citep{bu2025univla} & 7B & 1.1 & 52.6 & 83.9 & 96.6 & 90.7 & 15.7 & 69.5 & 55.5 \\
$\pi_0$ \citep{black2024pi0} & 3.3B & 17.8 & 6.6 & 58.8 & 89.7 & 90.7 & 90.9 & 89.1 & 60.7 \\
$\pi_0$-Fast \citep{pertsch2025fast} & 3.3B & 87.2 & 26.9 & 84.2 & 37.0 & 97.7 & 93.2 & 95.5 & 74.4 \\
RIPT-VLA \citep{tan2025riptvla} & 7B & 85.4 & 38.0 & \textbf{99.7} & 99.7 & \textbf{100.0} & 92.0 & 92.3 & 85.8 \\
OpenVLA-OFT+ \citep{fei2025liberoplus} & 7B & \textbf{98.4} & 31.7 & 96.0 & 99.3 & 98.8 & 86.3 & \textbf{97.8} & 86.1 \\
\midrule
\textbf{CoTinyVLA (Ours)} & \textbf{0.9B} & 97.3 & \textbf{58.3} & 96.4 & 98.6 & 99.2 & \textbf{98.0} & 90.1 & \textbf{90.8} \\
\bottomrule
\end{tabular}
\caption{\textbf{LIBERO-Plus Spatial} per-perturbation and overall success rate (\%) across 2{,}402 perturbed tasks. CoTinyVLA reaches the highest \emph{Total} success rate, 4.7 points above the strongest 7B baseline, and leads on \emph{Robot Initial States}, the axis on which no baseline exceeds 52.6\%, despite using approximately one-eighth the parameters. Bold values mark the best entry per column.}
\label{tab:lp-spatial}
\end{table*}

\section{Experiments}
\label{sec:experiments}

\subsection{Benchmarks}

\textbf{LIBERO-Plus} \citep{fei2025liberoplus} systematically perturbs the original LIBERO benchmark \citep{liu2023libero} along seven independent dimensions (Camera Viewpoints, Robot Initial States, Language Instructions, Light Conditions, Background Textures, Sensor Noise, Objects Layout), each at five difficulty levels. The benchmark covers 10{,}030 perturbed tasks across the four standard suites Spatial, Object, Goal, and Long; following the standard LIBERO-Plus protocol, each task is evaluated with a single rollout. Because the perturbed tasks are systematic modifications of the same 40 base tasks, each base task is covered by roughly 250 instances, so a suite total already aggregates hundreds of rollouts per underlying task and a single rollout per instance is sufficient. Under a binomial approximation, with $\sim$2{,}500 tasks per suite, the suite-level success rate has a standard error below one percentage point. Confidence intervals quoted below are reconstructed from published aggregates and are therefore approximate; the supplementary material reports them for all models.

\textbf{Standard LIBERO} is the original, unperturbed benchmark, used to verify that our model retains in-distribution capability. We follow the canonical protocol of 50 trials per task across 10 tasks per suite (500 trials per suite).

\textbf{Inference budget.} Following the OpenVLA-OFT family \citep{kim2025openvlaoft}, we cap each rollout at 220 environment steps on Spatial, 280 on Object, 300 on Goal and 520 on Long, on both benchmarks.

\subsection{Baselines}

We compare against all baselines reported in the LIBERO-Plus paper: OpenVLA \citep{kim2025openvla}, OpenVLA-OFT \citep{kim2025openvlaoft}, OpenVLA-OFT$_w$, OpenVLA-OFT$_m$, NORA \citep{hung2025nora}, WorldVLA \citep{cen2025worldvla}, UniVLA \citep{bu2025univla}, $\pi_0$ \citep{black2024pi0}, $\pi_0$-Fast \citep{pertsch2025fast}, RIPT-VLA \citep{tan2025riptvla}, and OpenVLA-OFT+ \citep{fei2025liberoplus}. Standard LIBERO numbers for these baselines are taken from their original publications. All baselines are 3B--7B in size.

\subsection{Main Result on LIBERO-Plus Spatial}
\label{sec:results-spatial}

Table~\ref{tab:lp-spatial} shows the per-perturbation success rates on LIBERO-Plus Spatial. CoTinyVLA reaches 90.8\% overall, the highest value on the suite and 4.7 points above OpenVLA-OFT+, which is seven times its size; the interval for the difference, $[+2.9, +6.5]$, excludes zero. It leads on Robot Initial States (58.3\% versus 52.6\% for the best 7B model, UniVLA \citep{bu2025univla}), the axis on which no baseline exceeds 52.6\% while every other axis has a baseline above 96\%, and on Sensor Noise (98.0\%). It also reaches 98.6\% on Light Conditions and 99.2\% on Background Textures and stays above 90\% on every axis except Robot Initial States.

\subsection{Results on Object, Goal, and Long}
\label{sec:results-others}

Table~\ref{tab:lp-object} reports the Object suite; per-perturbation results for the Goal and Long suites are reported in the supplementary material. CoTinyVLA reaches 87.3\% on Object, 86.6\% on Goal and 80.7\% on the long-horizon Long suite, 2.8, 15.9 and 3.0 points above the strongest 7B baseline, OpenVLA-OFT+, with every interval for the difference excluding zero. Goal carries the largest margin of the four suites: CoTinyVLA leads on all seven perturbation axes there, and by 33.7 and 27.1 points on Robot Initial States and Language Instructions, the two axes on which the strongest baseline scores lowest.

\begin{table*}[t]
\centering
\fontsize{9}{10.5}\selectfont
\setlength{\tabcolsep}{3.5pt}
\renewcommand{\arraystretch}{1.05}
\begin{tabular}{l r r r r r r r r r}
\toprule
\textbf{Model} & \textbf{Params} & \textbf{Camera} & \textbf{Robot} & \textbf{Lang.} & \textbf{Light} & \textbf{Backgr.} & \textbf{Noise} & \textbf{Layout} & \textbf{Total} \\
\midrule
OpenVLA \citep{kim2025openvla} & 7B & 0.5 & 4.5 & 21.0 & 1.0 & 45.2 & 11.4 & 22.4 & 14.0 \\
OpenVLA-OFT \citep{kim2025openvlaoft} & 7B & 38.9 & 25.4 & 99.0 & 73.7 & 97.6 & 72.3 & 71.8 & 66.5 \\
OpenVLA-OFT$_w$ \citep{kim2025openvlaoft} & 7B & 10.1 & 31.4 & 76.4 & 85.9 & 96.4 & 48.3 & 66.3 & 56.0 \\
OpenVLA-OFT$_m$ \citep{kim2025openvlaoft} & 7B & 70.2 & 18.1 & 98.5 & \textbf{100.0} & 91.9 & 94.1 & 77.4 & 77.1 \\
NORA \citep{hung2025nora} & 3B & 0.5 & 28.4 & 76.4 & 25.3 & 54.8 & 5.7 & 55.8 & 34.4 \\
WorldVLA \citep{cen2025worldvla} & 7B & 0.0 & 26.4 & 57.2 & 20.5 & 17.3 & 18.0 & 53.6 & 28.6 \\
UniVLA \citep{bu2025univla} & 7B & 0.0 & 42.2 & 86.9 & 25.6 & 81.5 & 10.4 & 27.3 & 36.7 \\
$\pi_0$ \citep{black2024pi0} & 3.3B & 22.2 & 8.3 & 70.0 & 90.9 & 91.1 & 87.0 & 76.2 & 61.4 \\
$\pi_0$-Fast \citep{pertsch2025fast} & 3.3B & 72.0 & 27.6 & 71.5 & 71.0 & 95.2 & 93.1 & 84.5 & 72.7 \\
RIPT-VLA \citep{tan2025riptvla} & 7B & 37.9 & 26.4 & 80.8 & 85.9 & 99.2 & 68.0 & 70.1 & 64.3 \\
OpenVLA-OFT+ \citep{fei2025liberoplus} & 7B & 97.0 & 24.6 & \textbf{100.0} & 99.7 & 98.8 & 97.4 & 82.8 & 84.5 \\
\midrule
\textbf{CoTinyVLA (Ours)} & \textbf{0.9B} & \textbf{99.7} & \textbf{44.0} & 89.8 & \textbf{100.0} & \textbf{100.0} & \textbf{99.5} & \textbf{85.6} & \textbf{87.3} \\
\bottomrule
\end{tabular}
\caption{\textbf{LIBERO-Plus Object} per-perturbation and overall success rate (\%) across 2{,}518 perturbed tasks. CoTinyVLA leads on \emph{Total} and on five of the seven perturbation axes, and is the only model to reach 100\% on both \emph{Light Conditions} and \emph{Background Textures}. Bold values mark the best entry per column.}
\label{tab:lp-object}
\end{table*}

\subsection{Results on Standard LIBERO}
\label{sec:results-standard}

On standard, unperturbed LIBERO (Table~\ref{tab:standard}), the four suites give 99.4\%, 100.0\%, 98.6\% and 92.0\%, averaging 97.5\%. This matches the strongest 7B baseline (RIPT-VLA 97.5\%) and is ahead of OpenVLA-OFT (97.1\%).

\begin{table*}[!t]
\centering
\fontsize{9}{10.5}\selectfont
\setlength{\tabcolsep}{10pt}
\renewcommand{\arraystretch}{1.02}
\begin{tabular}{l r r r r r r}
\toprule
\textbf{Model} & \textbf{Pars.} & \textbf{Spat.} & \textbf{Obj.} & \textbf{Goal} & \textbf{Long} & \textbf{Avg.} \\
\midrule
\multicolumn{7}{l}{\emph{Sub-1B models}} \\
\midrule
HyperVLA \citep{xiong2025hypervla}        & 86.1M & 95.0 & 94.0 & 92.0 & 74.0 & 89.0 \\
NanoVLA-S \citep{chen2025nanovla}         & 161M  & 81.6 & 93.6 & 89.6 & 49.8 & 78.7 \\
SmolVLA \citep{shukor2025smolvla}         & 0.24B & 87.0 & 93.0 & 88.0 & 63.0 & 82.8 \\
NanoVLA-R                                 & 296M  & 89.8 & 96.2 & 93.0 & 57.4 & 84.1 \\
CorridorVLA \citep{li2026corridorvla}     & 0.45B & 92.0 & 95.8 & 90.8 & 85.2 & 91.0 \\
SmolVLA                                   & 0.45B & 90.0 & 96.0 & 92.0 & 71.0 & 87.3 \\
SwiftVLA \citep{ni2025swiftvla}           & 0.45B & \textit{97.0} & 96.4 & \textit{96.8} & 88.4 & 94.7 \\
RD-VLA (Adaptive) \citep{tur2026rdvla}    & 0.5B  & 88.6 & 98.8 & \textit{96.8} & 85.8 & 92.5 \\
RD-VLA (Fixed)                            & 0.5B  & 92.0 & \textit{99.0} & 96.0 & 84.8 & 93.0 \\
UniAct \citep{zheng2025uniact}            & 0.5B  & 77.0 & 87.0 & 77.0 & 70.0 & 77.8 \\
VLA-0-Smol \citep{balakhnov2025vla0smol}  & 0.5B  & 92.2 & 97.2 & 95.6 & 91.2 & 94.1 \\
VLA-OS \citep{gao2025vlaos}               & 0.5B  & 87.0 & 96.5 & 92.7 & 66.0 & 85.6 \\
NanoVLA-L                                 & 520M  & 87.2 & 89.8 & 90.0 & 55.2 & 80.4 \\
Evo-1 \citep{lin2025evo1}                 & 0.77B & 92.7 & 97.7 & 96.3 & \textbf{92.3} & \textit{94.8} \\
InspireVLA \citep{zhang2025inspire}       & 1B    & 90.7 & 94.3 & 88.3 & 73.3 & 86.7 \\
\midrule
\textbf{CoTinyVLA (Ours)}                 & 0.9B  & \textbf{99.4} & \textbf{100.0} & \textbf{98.6} & \textit{92.0} & \textbf{97.5} \\
\midrule
\multicolumn{7}{l}{\emph{Models above 1B}} \\
\midrule
SwiftVLA + 4D                             & 1.65B & 97.2 & 96.8 & 97.4 & 89.0 & 95.1 \\
GraspVLA \citep{deng2025graspvla}         & 1.8B  & --  & 94.1 & 91.2 & 82.0 & 89.1 \\
SmolVLA                                   & 2.25B & 93.0 & 94.0 & 91.0 & 77.0 & 88.8 \\
GR00T-N1 \citep{bjorck2025grootn1}        & 2--3B & 94.4 & 97.6 & 93.0 & 90.6 & 93.9 \\
NORA \citep{hung2025nora}                 & 3B    & 92.2 & 95.4 & 89.4 & 74.6 & 87.9 \\
NORA-1.5 \citep{hung2025nora15}           & 3B    & 98.0 & 96.0 & 95.4 & 90.5 & 95.0 \\
VLA-0 \citep{goyal2025vla0}               & 3B    & 97.8 & 97.0 & 96.2 & 87.6 & 94.7 \\
$\pi_{0.5}$ \citep{black2025pi05}         & 3B    & \textit{98.8} & 98.2 & 98.0 & 92.4 & 96.9 \\
$\pi_0$ \citep{black2024pi0}              & 3.3B  & 96.8 & \textit{98.8} & 95.8 & 85.2 & 94.2 \\
$\pi_0$-Fast \citep{pertsch2025fast}      & 3.3B  & 96.4 & 96.8 & 88.6 & 60.2 & 85.5 \\
4D-VLA \citep{zhang2025fourdvla}          & 4B    & 88.9 & 95.2 & 90.9 & 79.1 & 88.6 \\
QDepth-VLA \citep{li2025qdepthvla}        & 4B    & 97.6 & 96.6 & 95.2 & 90.0 & 94.9 \\
SpatialVLA \citep{qu2025spatialvla}       & 4B    & 88.2 & 89.9 & 78.6 & 55.5 & 78.1 \\
OpenVLA \citep{kim2025openvla}            & 7B    & 84.7 & 88.4 & 79.2 & 53.7 & 76.5 \\
OpenVLA-OFT \citep{kim2025openvlaoft}     & 7B    & 97.6 & 98.4 & 97.9 & \textbf{94.5} & \textit{97.1} \\
RIPT-VLA \citep{tan2025riptvla}           & 7B    & 98.6 & 98.6 & \textbf{99.0} & \textit{93.8} & \textbf{97.5} \\
UniVLA \citep{bu2025univla}               & 7B    & 96.5 & 96.8 & 95.6 & 92.0 & 95.2 \\
WorldVLA \citep{cen2025worldvla}          & 7B    & 85.6 & 89.0 & 82.6 & 59.0 & 79.1 \\
\midrule
\textbf{CoTinyVLA (Ours)}                 & 0.9B  & \textbf{99.4} & \textbf{100.0} & \textit{98.6} & 92.0 & \textbf{97.5} \\
\bottomrule
\end{tabular}
\caption{\textbf{Standard LIBERO} (unperturbed) 4-suite success rate (\%): 50 trials per task, 10 tasks per suite. Baselines are sorted by parameter count. CoTinyVLA is listed twice, once at the end of the sub-1B block and once at the end of the full table, so that it can be read against either comparison set. Within the sub-1B block, \textbf{bold} and \textit{italic} mark the best and second-best entry per column among sub-1B models; in the remainder of the table they mark the best and second-best entry per column among the models above 1B together with CoTinyVLA. CoTinyVLA averages 97.5\% across the four suites, matching the strongest reported 7B baseline (RIPT-VLA 97.5\%) and 2.7 points above the strongest reported sub-1B baseline (Evo-1 94.8\%).}
\label{tab:standard}
\end{table*}

\section{Analysis}
\label{sec:analysis}

\subsection{Per-Perturbation Analysis}

Robot Initial States, which injects kinematic offsets in the starting pose, is the hardest axis of the benchmark: no baseline exceeds 53.2\% on it in any suite, against 80\% to 100\% on the visual axes. It is also where CoTinyVLA gains most, by $+5.7$ points over the best baseline on Spatial and $+33.7$ on Goal. Doing well under it requires the model to locate the gripper and the target object accurately and then adapt the trajectory. Dual-view temporal input gives the model 16 frames per step, providing temporal observations from which approach velocity and orientation can be inferred in both views. Our input ablation isolates this effect directly: within that sweep, replacing the eight-frame history with a single frame while keeping both views costs 16.0 points on this axis. A test-time intervention points the same way: repeating the most recent frame across the history window, at unchanged token count, takes standard-LIBERO success from 100\% to 60\% (supplementary material). The \textsc{Think} slot \textsc{Phase} supplies an additional discrete signal aligned with kinematic phase, and its stability tracks the perturbations that hurt: on Spatial the regression rate rises from 0.8\% to 15.2\% when the history is replaced by a repeated frame.

\subsection{Per-Perturbation Pattern Across Suites}

Across all four LIBERO-Plus suites the per-perturbation profile is consistent, and the supplementary material gives a per-suite radar comparison against every baseline. The model stays above 90\% on the visual perturbations (Light Conditions, Background Textures, Sensor Noise) in all four suites and above 96\% in three of them. Linguistic perturbation (Language Instructions) holds above 68\% on every suite; the Spatial-suite ablation attributes that robustness to the paraphrase augmentation. Robot Initial States and Objects Layout are the hardest dimensions, at 57.0\% and 78.4\% aggregated over the benchmark; both involve physical-state shifts.

Each perturbation is further divided into five difficulty levels, and the same split holds there. On Spatial, Light and Background stay at or above 93\% from Level~1 to Level~5 and Objects Layout falls from 100\% to 84\%, while Robot Initial States degrades from 94\% to 7\%, so one dimension carries the suite's difficulty scaling. The supplementary material plots all seven.

\section{Discussion}
\label{sec:discussion}

The results indicate that on LIBERO-Plus the robustness usually attributed to 3--7B backbones does not require them: structuring supervision along the axes the benchmark perturbs is enough to exceed those systems at 0.9B. Several boundaries limit how far that conclusion carries.

The evaluation is in simulation on a single embodiment, LIBERO's Franka Panda, so sim-to-real transfer of the dual-view temporal input and cross-embodiment transfer are untested, as are history lengths beyond eight frames per view. Inference cost is measured in a synchronous implementation on a single L40S; decoding accounts for most of the 1.37\,s per chunk, so the cost is set by generation and not by the architecture, and \textsc{Plan} caching already halves it, but overlapping generation with execution and end-to-end latency on a physical robot are not evaluated.

The method also carries dependencies of its own. Distillation needs a strong vision-language teacher, and cheaper teachers may degrade label quality, particularly for the structured \textsc{Think} attributes; we characterise the generated paraphrases by lexical statistics, and establishing their semantic equivalence to the base instructions would need a separate evaluation. One weakness survives the gains reported above. The Long suite is the lowest of the four in absolute terms at 80.7\%, and its Language Instructions axis carries the largest per-axis deficit anywhere in the benchmark, 68.9\% against 94.8\% for the strongest baseline. A compact backbone with an eight-frame history is therefore still at a disadvantage when long-horizon execution and instruction variation combine.

Evaluation variance is characterised: each suite total aggregates roughly 250 perturbed instances per base task, per-suite and axis-level intervals are reported in the supplementary material, and the test-time interventions are paired within a single trained model and tested exactly, so the mechanism results do not depend on the training seed. Variance across training runs is the one source that remains uncharacterised; evaluating additional seeds and other multimodal backbone families is left for future work.

\section{Conclusion}
\label{sec:conclusion}

We have presented CoTinyVLA, a 0.9B-parameter vision-language-action model that leads the strongest 7B baselines on all four LIBERO-Plus suites, by 4.7 to 15.9 points with every margin interval excluding zero, at roughly one-eighth of the parameter count and within a 2.25\,GiB inference footprint. Our ablations separate the three components by axis: the visual input configuration accounts for the largest total-score reduction, hierarchical CoT distillation acts on the physical-state perturbations, and paraphrase augmentation moves the Language Instructions axis by 12.3 points and leaves the rest unchanged. How a fixed image budget is divided between the two cameras and across time is itself worth 8.6 points, so input structure is a design lever, not an overhead the architecture simply pays.


\bibliographystyle{plainnat}
\bibliography{custom}


\clearpage
\appendix

\section{Qualitative Rollouts with CoT Outputs}
\label{app:rollouts}

\begin{figure*}[!t]
\centering
\includegraphics[width=0.8\textwidth]{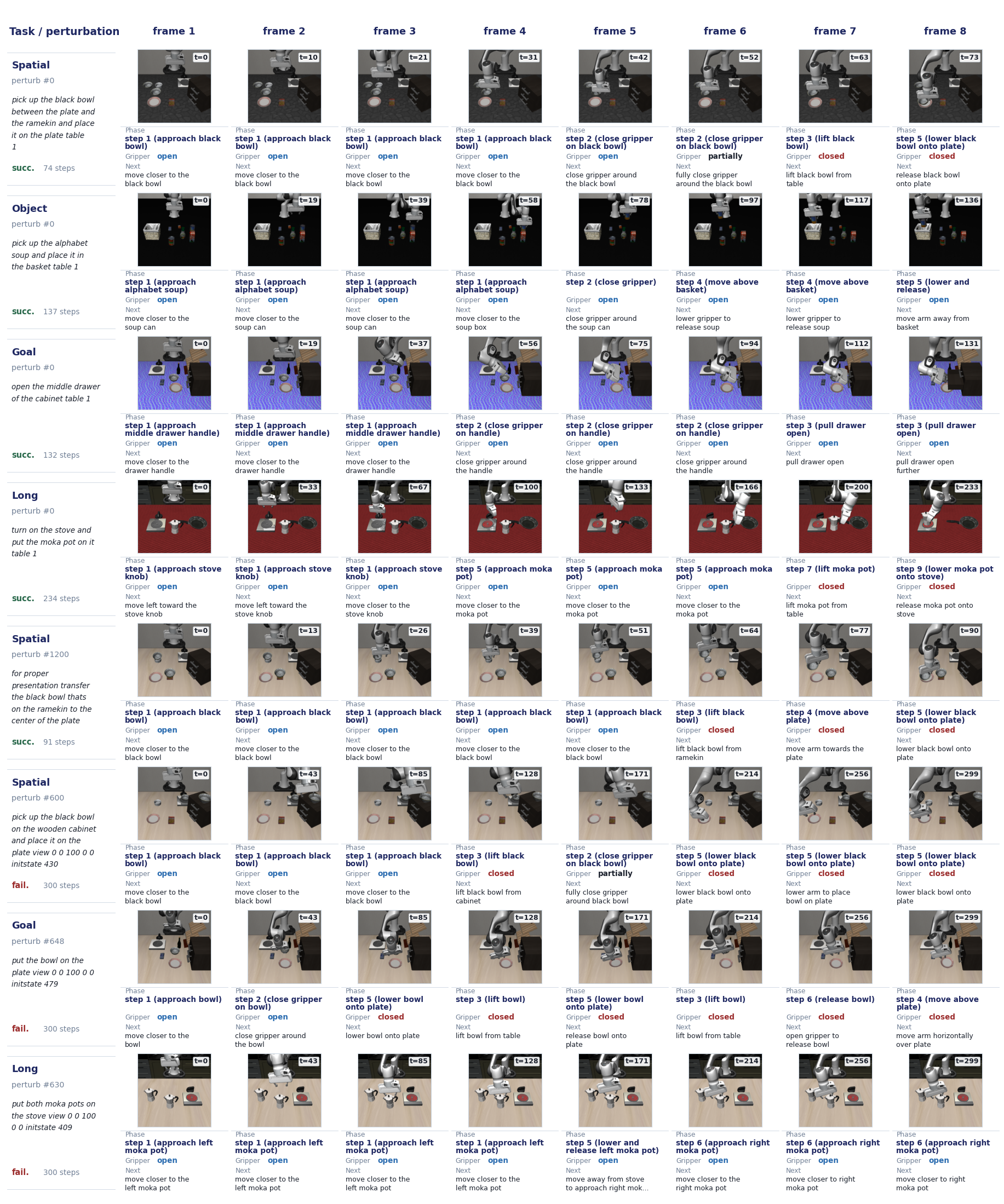}
\caption{Eight LIBERO-Plus rollouts with CoTinyVLA's emitted CoT tokens shown beneath each frame. Top four rows: successful unperturbed rollouts (Spatial, Object, Goal, Long, perturbation 0). Fifth row: successful rollout under Spatial perturbation 1200. Bottom three rows: failures under perturbations 600 (Spatial), 648 (Goal) and 630 (Long), each hitting the step budget of its suite (220, 300 and 520 respectively). For each chunk the structured \textsc{Think} output is shown: \textsc{Phase} (current subaction step), \textsc{Gripper} state, and \textsc{Next} (a short natural-language continuation).}
\label{fig:rollouts}
\end{figure*}

Figure~\ref{fig:rollouts} shows eight LIBERO-Plus rollouts from CoTinyVLA, with the model's emitted \textsc{Think} tokens (Phase, Gripper, Next) printed beneath each rendered frame. The top four rows are successful rollouts on the unperturbed instances of the four LIBERO-Plus suites (Spatial, Object, Goal, Long). The fifth row is a successful rollout under a Spatial perturbation. The bottom three rows are failures on perturbed Spatial, Goal, and Long instances, each of which exhausts the inference budget of its suite.

Beneath each frame we show the structured \textsc{Think} output for the corresponding action chunk: the predicted \textsc{Phase} (the current subaction stage, e.g., ``step~3 (lift black bowl)''), the \textsc{Gripper} state (\textsc{OPEN}, \textsc{PARTIALLY\_CLOSED}, or \textsc{CLOSED}), and a short natural-language \textsc{Next} descriptor. The \textsc{Think} tokens are produced before the corresponding action chunk, so they are part of the model's prediction for that chunk.

In the successful rollouts, \textsc{Phase} transitions track the task progression coherently (approach $\to$ close gripper $\to$ lift $\to$ move above target $\to$ lower $\to$ release), and the \textsc{Gripper} state switches monotonically from \textsc{OPEN} through \textsc{PARTIALLY\_CLOSED} to \textsc{CLOSED} as the grasp completes. In the failed rollouts, \textsc{Phase} predictions oscillate or regress. The failed Spatial rollout (perturbation 600) jumps between step~3 (lift), step~2 (close gripper), and step~5 (lower onto plate) within a few chunks, and the failed Goal rollout (perturbation 648) alternates among step~3, step~5, and step~6 without converging. Many failed frames also show a mismatch between the predicted \textsc{Phase} and the actual scene state. In Spatial perturbation 600 (frames 4--8), the predicted \textsc{Phase} advances to a later stage of the task (``lift black bowl'', ``lower onto plate'') while the gripper is still empty and the target object has not moved; the Long rollout (perturbation 630) shows the same pattern, with the predicted \textsc{Phase} advancing to the second moka pot before the first is grasped. These examples illustrate a failure mode that combines planning misalignment, made visible by the \textsc{Think} schema, with incorrect inference about the current visual state.

\section{Per-Suite Radar Comparison}
\label{app:radar}

\begin{figure*}[!t]
\centering
\input{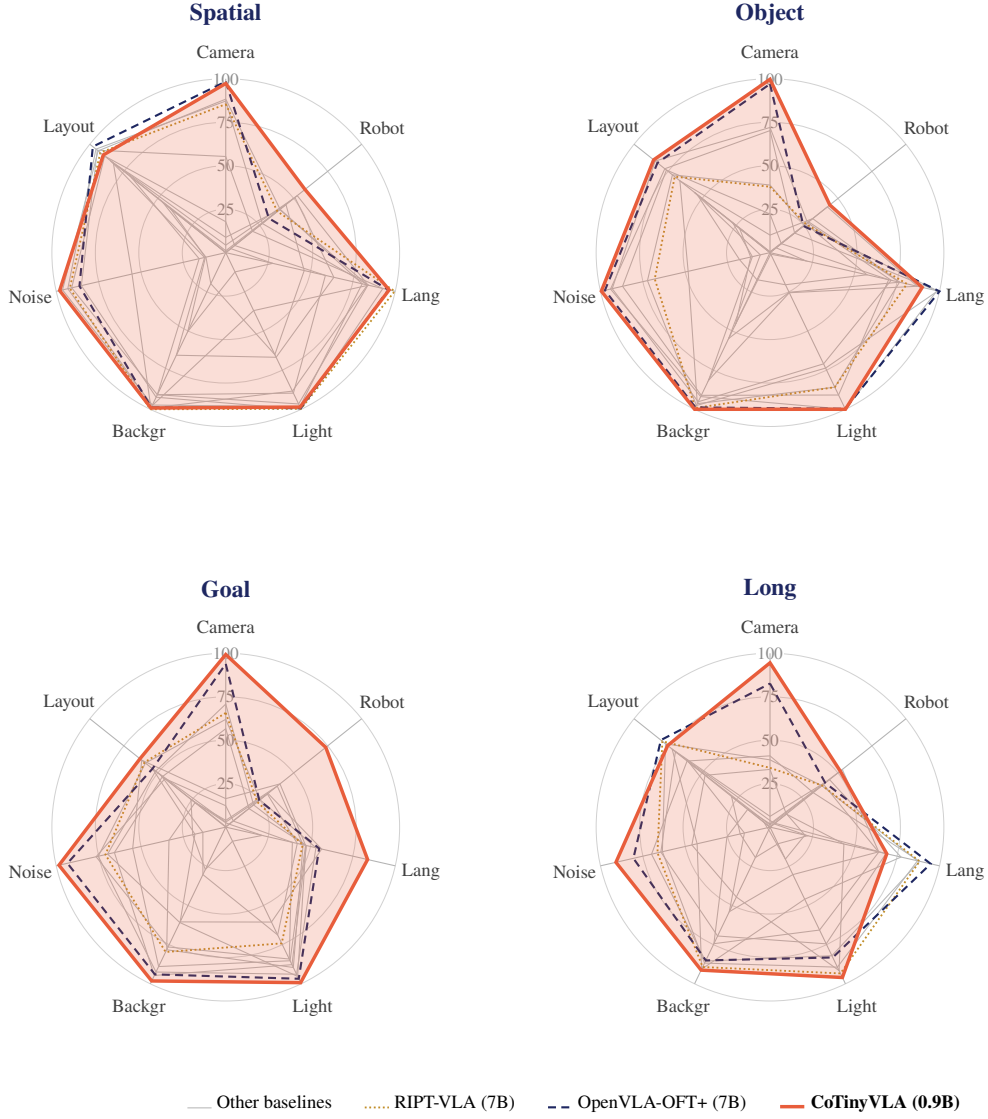}
\caption{Per-suite radar comparison of all LIBERO-Plus baselines and CoTinyVLA, with one panel per suite (Spatial, Object, Goal, Long). The remaining nine baselines (OpenVLA, OpenVLA-OFT, OpenVLA-OFT$_w$, OpenVLA-OFT$_m$, NORA, WorldVLA, UniVLA, $\pi_0$, $\pi_0$-Fast) are drawn as thin grey polygons; the strongest baseline overall, OpenVLA-OFT+, and the strongest on the Spatial suite after it, RIPT-VLA, are emphasised in dotted gold and dashed navy; CoTinyVLA (0.9B) is drawn in solid coral with light shading.}
\label{fig:radar}
\end{figure*}

Figure~\ref{fig:radar} compares CoTinyVLA (highlighted in coral) against every LIBERO-Plus baseline, with one radar plot per suite (Spatial, Object, Goal, Long). Each radar has seven axes, one per perturbation dimension, and each model is drawn as a closed polygon over those axes. CoTinyVLA reaches high values across most axes on Spatial and Object, is competitive with the strongest 7B baseline on Goal, and shows lower scores on several axes in the Long suite.

\section{LIBERO-Plus Goal Suite Results}
\label{app:goal}

Table~\ref{tab:lp-goal} gives the full per-perturbation breakdown for LIBERO-Plus Goal, the suite on which CoTinyVLA holds its largest margin. It leads on all seven axes, and the two axes on which the baselines are weakest, Robot Initial States and Language Instructions, are the two on which it gains most.

\begin{table*}[t]
\centering
\fontsize{9}{10.5}\selectfont
\setlength{\tabcolsep}{3.5pt}
\renewcommand{\arraystretch}{1.05}
\begin{tabular}{l r r r r r r r r r}
\toprule
\textbf{Model} & \textbf{Params} & \textbf{Camera} & \textbf{Robot} & \textbf{Lang.} & \textbf{Light} & \textbf{Backgr.} & \textbf{Noise} & \textbf{Layout} & \textbf{Total} \\
\midrule
OpenVLA \citep{kim2025openvla} & 7B & 2.5 & 2.7 & 21.5 & 9.0 & 27.1 & 19.5 & 25.6 & 15.1 \\
OpenVLA-OFT \citep{kim2025openvlaoft} & 7B & 62.0 & 25.2 & 53.2 & 93.9 & 92.5 & 75.2 & 59.1 & 63.0 \\
OpenVLA-OFT$_w$ \citep{kim2025openvlaoft} & 7B & 16.4 & 39.9 & 47.1 & 85.3 & 89.0 & 54.9 & 61.8 & 53.3 \\
OpenVLA-OFT$_m$ \citep{kim2025openvlaoft} & 7B & 56.6 & 17.1 & 47.6 & 87.8 & 94.7 & 65.7 & 46.6 & 56.2 \\
NORA \citep{hung2025nora} & 3B & 2.9 & 31.1 & 56.6 & 60.6 & 60.5 & 18.2 & 53.9 & 38.8 \\
WorldVLA \citep{cen2025worldvla} & 7B & 0.3 & 30.6 & 42.2 & 68.8 & 30.3 & 13.5 & 47.4 & 31.8 \\
UniVLA \citep{bu2025univla} & 7B & 3.9 & 37.9 & 45.6 & 89.6 & 78.3 & 33.5 & 22.6 & 40.7 \\
$\pi_0$ \citep{black2024pi0} & 3.3B & 12.3 & 5.6 & 39.3 & 84.2 & 76.5 & 76.5 & 44.7 & 44.9 \\
$\pi_0$-Fast \citep{pertsch2025fast} & 3.3B & 70.8 & 20.5 & 47.3 & 95.3 & 60.9 & 69.7 & 51.6 & 57.5 \\
RIPT-VLA \citep{tan2025riptvla} & 7B & 65.7 & 23.2 & 45.4 & 74.2 & 79.7 & 71.0 & 59.8 & 58.0 \\
OpenVLA-OFT+ \citep{fei2025liberoplus} & 7B & 93.9 & 24.7 & 55.1 & 96.8 & 94.0 & 93.4 & 53.9 & 70.7 \\
\midrule
\textbf{CoTinyVLA (Ours)} & \textbf{0.9B} & \textbf{99.3} & \textbf{73.6} & \textbf{83.7} & \textbf{99.3} & \textbf{98.2} & \textbf{98.4} & \textbf{63.1} & \textbf{86.6} \\
\bottomrule
\end{tabular}
\caption{\textbf{LIBERO-Plus Goal} per-perturbation and overall success rate (\%) across 2{,}591 perturbed tasks. CoTinyVLA leads on \emph{Total} by 15.9 points and on every perturbation axis, with the largest margins on \emph{Robot Initial States} and \emph{Language Instructions}. Bold values mark the best entry per column.}
\label{tab:lp-goal}
\end{table*}

\section{LIBERO-Plus Long Suite Results}
\label{app:long}

Table~\ref{tab:lp-long} gives the full per-perturbation breakdown for the long-horizon Long suite. CoTinyVLA leads the total and four of the seven axes; it trails on Language Instructions, Objects Layout and, by 1.5 points, Robot Initial States, the three axes where accumulated execution error over a longer horizon compounds with the perturbation.

\begin{table*}[!t]
\centering
\fontsize{9}{10.5}\selectfont
\setlength{\tabcolsep}{3.5pt}
\renewcommand{\arraystretch}{1.05}
\begin{tabular}{l r r r r r r r r r}
\toprule
\textbf{Model} & \textbf{Params} & \textbf{Camera} & \textbf{Robot} & \textbf{Lang.} & \textbf{Light} & \textbf{Backgr.} & \textbf{Noise} & \textbf{Layout} & \textbf{Total} \\
\midrule
OpenVLA \citep{kim2025openvla} & 7B & 0.0 & 3.0 & 22.2 & 10.6 & 19.4 & 17.6 & 28.3 & 14.3 \\
OpenVLA-OFT \citep{kim2025openvlaoft} & 7B & 38.7 & 38.2 & 87.0 & 89.4 & 86.8 & 63.5 & 76.9 & 66.4 \\
OpenVLA-OFT$_w$ \citep{kim2025openvlaoft} & 7B & 6.2 & 43.8 & 77.3 & 46.0 & 90.7 & 43.0 & 72.0 & 52.2 \\
OpenVLA-OFT$_m$ \citep{kim2025openvlaoft} & 7B & 41.0 & 31.8 & 88.3 & 82.1 & 85.5 & 69.9 & 61.0 & 63.9 \\
NORA \citep{hung2025nora} & 3B & 1.2 & 39.4 & 64.0 & 30.3 & 54.0 & 15.1 & 59.5 & 36.3 \\
WorldVLA \citep{cen2025worldvla} & 7B & 0.0 & 12.2 & 20.6 & 20.4 & 1.7 & 1.6 & 4.4 & 8.2 \\
UniVLA \citep{bu2025univla} & 7B & 1.9 & \textbf{53.2} & 64.2 & 65.7 & 74.4 & 25.4 & 16.4 & 39.9 \\
$\pi_0$ \citep{black2024pi0} & 3.3B & 3.8 & 3.6 & 68.4 & 74.5 & 69.5 & 64.4 & 69.6 & 48.4 \\
$\pi_0$-Fast \citep{pertsch2025fast} & 3.3B & 33.2 & 12.0 & 43.6 & 91.6 & 44.6 & 46.1 & 47.8 & 43.4 \\
RIPT-VLA \citep{tan2025riptvla} & 7B & 34.1 & 38.4 & 88.3 & 93.4 & 89.3 & 66.4 & 79.2 & 67.5 \\
OpenVLA-OFT+ \citep{fei2025liberoplus} & 7B & 82.6 & 40.7 & \textbf{94.8} & 83.2 & 85.1 & 80.6 & \textbf{80.3} & 77.7 \\
\midrule
\textbf{CoTinyVLA (Ours)} & \textbf{0.9B} & \textbf{94.5} & 51.7 & 68.9 & \textbf{96.0} & \textbf{91.3} & \textbf{90.9} & 75.3 & \textbf{80.7} \\
\bottomrule
\end{tabular}
\caption{\textbf{LIBERO-Plus Long} per-perturbation and overall success rate (\%) across 2{,}519 perturbed tasks. CoTinyVLA reaches the highest \emph{Total} success rate on the long-horizon suite, 3.0 points above the strongest 7B baseline, while trailing it on \emph{Language Instructions} and \emph{Objects Layout}. Bold values mark the best entry per column.}
\label{tab:lp-long}
\end{table*}

\section{Per-Suite Difficulty Breakdown}
\label{app:difficulty}

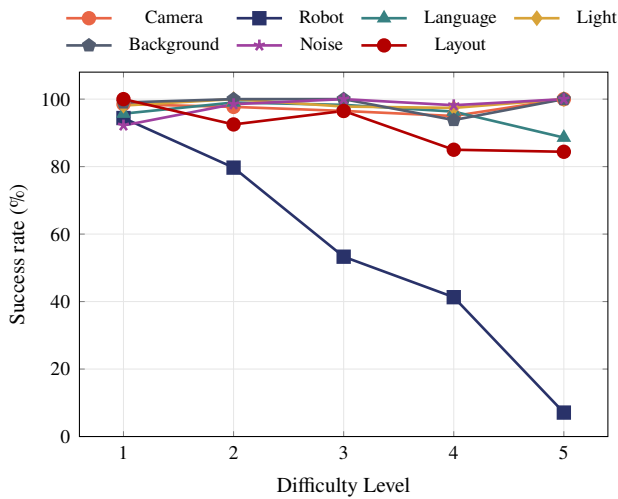
\begin{figure}[!t]
\centering
\begin{tikzpicture}
\begin{axis}[
    width=\columnwidth,
    height=6.4cm,
    xlabel={Difficulty Level},
    ylabel={Success rate (\%)},
    xlabel style={font=\footnotesize},
    ylabel style={font=\footnotesize, yshift=-2pt},
    xtick={1,2,3,4,5},
    ymin=0, ymax=108,
    ytick={0,20,40,60,80,100},
    xticklabel style={font=\scriptsize},
    yticklabel style={font=\scriptsize},
    legend style={
        font=\scriptsize,
        at={(0.5,1.02)}, anchor=south,
        legend columns=4,
        column sep=4pt,
        draw=none,
        fill=none,
    },
    grid=major,
    grid style={line width=0.2pt, draw=black!10},
    axis line style={line width=0.4pt},
    tick style={black!50},
    major tick length=2pt,
    every axis plot/.append style={line width=1.0pt, mark size=2.2pt},
    cycle list name=color list,
]

\addplot[color=coral!95, mark=*, mark options={fill=coral!95}] coordinates {(1,98.5) (2,97.7) (3,96.5) (4,95.0) (5,100.0)};
\addlegendentry{Camera}

\addplot[color=navy!95, mark=square*, mark options={fill=navy!95}] coordinates {(1,94.4) (2,79.7) (3,53.3) (4,41.3) (5,7.1)};
\addlegendentry{Robot}

\addplot[color=teal!95, mark=triangle*, mark options={fill=teal!95}] coordinates {(1,95.7) (2,99.0) (3,98.3) (4,96.3) (5,88.6)};
\addlegendentry{Language}

\addplot[color=gold!95, mark=diamond*, mark options={fill=gold!95}] coordinates {(1,98.1) (2,100.0) (3,97.8) (4,97.4) (5,100.0)};
\addlegendentry{Light}

\addplot[color=slate!95, mark=pentagon*, mark options={fill=slate!95}] coordinates {(1,99.0) (2,100.0) (3,100.0) (4,93.8) (5,100.0)};
\addlegendentry{Background}

\addplot[color=violet!75, mark=star, mark options={fill=violet!75, draw=violet!75}] coordinates {(1,92.2) (2,98.5) (3,100.0) (4,98.2) (5,100.0)};
\addlegendentry{Noise}

\addplot[color=red!70!black, mark=otimes*, mark options={fill=red!70!black, draw=red!70!black}] coordinates {(1,100.0) (2,92.5) (3,96.5) (4,85.0) (5,84.4)};
\addlegendentry{Layout}

\end{axis}
\end{tikzpicture}
\caption{LIBERO-Plus Spatial: CoTinyVLA success rate (\%) as a function of difficulty level (1--5), per perturbation dimension. Every axis except Robot Initial States stays above 84\% at the hardest level, so the suite's difficulty scaling is carried by that one dimension.}
\label{fig:difficulty}
\end{figure}

Figure~\ref{fig:difficulty} plots the success rate of CoTinyVLA on LIBERO-Plus Spatial as a function of difficulty level, and we report the full per-perturbation $\times$ per-difficulty success rate breakdown for our model in Tables~\ref{tab:diff-spatial}--\ref{tab:diff-long}. These tables disaggregate the totals reported in the main text and substantiate the difficulty-degradation discussion in \S\ref{sec:analysis}.

\begin{table}[t]
\centering
\fontsize{9}{10.5}\selectfont
\setlength{\tabcolsep}{3.5pt}
\renewcommand{\arraystretch}{1.12}
\begin{tabular}{l r r r r r r}
\toprule
\textbf{Perturb.} & \textbf{Lv1} & \textbf{Lv2} & \textbf{Lv3} & \textbf{Lv4} & \textbf{Lv5} & \textbf{All} \\
\midrule
Camera     & 98.5 & 97.7 & 96.5 & 95.0 & 100.0 & 97.3 \\
Robot      & 94.4 & 79.7 & 53.3 & 41.3 & 7.1 & 58.3 \\
Language   & 95.7 & 99.0 & 98.3 & 96.3 & 88.6 & 96.4 \\
Light      & 98.1 & 100.0 & 97.8 & 97.4 & 100.0 & 98.6 \\
Backgr.    & 99.0 & 100.0 & 100.0 & 93.8 & 100.0 & 99.2 \\
Noise      & 92.2 & 98.5 & 100.0 & 98.2 & 100.0 & 98.0 \\
Layout     & 100.0 & 92.5 & 96.5 & 85.0 & 84.4 & 90.1 \\
\bottomrule
\end{tabular}
\caption{\textbf{CoTinyVLA on LIBERO-Plus Spatial:} per-perturbation success rate (\%) at each difficulty level. Every axis except Robot Initial States stays above 84\% at the hardest level; Robot Initial States falls from 94.4\% to 7.1\%, so almost all of the suite's difficulty scaling is carried by that one dimension.}
\label{tab:diff-spatial}
\end{table}

\begin{table}[t]
\centering
\fontsize{9}{10.5}\selectfont
\setlength{\tabcolsep}{3.5pt}
\renewcommand{\arraystretch}{1.12}
\begin{tabular}{l r r r r r r}
\toprule
\textbf{Perturb.} & \textbf{Lv1} & \textbf{Lv2} & \textbf{Lv3} & \textbf{Lv4} & \textbf{Lv5} & \textbf{All} \\
\midrule
Camera     & 100.0 & 100.0 & 100.0 & 98.3 & 100.0 & 99.7 \\
Robot      & 96.0 & 72.9 & 48.9 & 21.8 & 12.9 & 44.0 \\
Language   & 89.3 & 89.6 & 87.5 & 92.1 & 91.8 & 89.8 \\
Light      & 100.0 & 100.0 & 100.0 & 100.0 & 100.0 & 100.0 \\
Backgr.    & 100.0 & 100.0 & 100.0 & 100.0 & 100.0 & 100.0 \\
Noise      & 100.0 & 100.0 & 98.6 & 99.0 & 100.0 & 99.5 \\
Layout     & 100.0 & 97.4 & 98.4 & 78.8 & 68.6 & 85.6 \\
\bottomrule
\end{tabular}
\caption{\textbf{CoTinyVLA on LIBERO-Plus Object:} per-perturbation success rate (\%) at each difficulty level. The visual axes stay above 98\% at every level and Language Instructions is flat, while the two physical-state axes fall sharply, Robot Initial States from 96.0\% to 12.9\% and Objects Layout from 100.0\% to 68.6\%.}
\label{tab:diff-object}
\end{table}

\begin{table}[t]
\centering
\fontsize{9}{10.5}\selectfont
\setlength{\tabcolsep}{3.5pt}
\renewcommand{\arraystretch}{1.12}
\begin{tabular}{l r r r r r r}
\toprule
\textbf{Perturb.} & \textbf{Lv1} & \textbf{Lv2} & \textbf{Lv3} & \textbf{Lv4} & \textbf{Lv5} & \textbf{All} \\
\midrule
Camera     & 100.0 & 99.1 & 99.0 & 100.0 & 98.7 & 99.3 \\
Robot      & 97.7 & 82.9 & 84.4 & 69.6 & 49.5 & 73.6 \\
Language   & 95.6 & 100.0 & 98.1 & 97.5 & 66.5 & 83.7 \\
Light      & 100.0 & 98.0 & 100.0 & 100.0 & 100.0 & 99.3 \\
Backgr.    & 100.0 & 97.1 & 98.6 & 100.0 & 95.6 & 98.2 \\
Noise      & 94.9 & 99.1 & 100.0 & 98.0 & 100.0 & 98.4 \\
Layout     & 93.1 & 83.6 & 58.3 & 21.6 & 25.0 & 63.1 \\
\bottomrule
\end{tabular}
\caption{\textbf{CoTinyVLA on LIBERO-Plus Goal:} per-perturbation success rate (\%) at each difficulty level. The visual axes stay above 94\% throughout. Objects Layout degrades most on this suite, from 93.1\% to 25.0\%, ahead of Robot Initial States, which remains at 49.5\% at the hardest level, higher than on any other suite.}
\label{tab:diff-goal}
\end{table}

\begin{table}[t]
\centering
\fontsize{9}{10.5}\selectfont
\setlength{\tabcolsep}{3.5pt}
\renewcommand{\arraystretch}{1.12}
\begin{tabular}{l r r r r r r}
\toprule
\textbf{Perturb.} & \textbf{Lv1} & \textbf{Lv2} & \textbf{Lv3} & \textbf{Lv4} & \textbf{Lv5} & \textbf{All} \\
\midrule
Camera     & 93.8 & 97.9 & 98.2 & 93.1 & 93.4 & 94.5 \\
Robot      & 92.2 & 74.0 & 53.0 & 31.3 & 11.6 & 51.7 \\
Language   & 78.5 & 84.4 & 72.7 & 69.2 & 42.5 & 68.9 \\
Light      & 100.0 & 100.0 & 100.0 & 97.3 & 91.2 & 96.0 \\
Backgr.    & 100.0 & 94.6 & 98.1 & 91.8 & 83.2 & 91.3 \\
Noise      & 100.0 & 93.8 & 93.8 & 91.9 & 83.3 & 90.9 \\
Layout     & 73.3 & 74.3 & 81.5 & 70.0 & 100.0 & 75.3 \\
\bottomrule
\end{tabular}
\caption{\textbf{CoTinyVLA on LIBERO-Plus Long:} per-perturbation success rate (\%) at each difficulty level. Long-horizon execution compounds with perturbation, so the visual axes also decline here, to 83--91\% at Level~5, and Language Instructions falls to 42.5\%. Robot Initial States again degrades most, from 92.2\% to 11.6\%. The Objects Layout row is non-monotonic and rests on few instances per level.}
\label{tab:diff-long}
\end{table}

\section{Confidence Intervals for the Reported Success Rates}
\label{app:ci}

Every LIBERO-Plus task is evaluated with a single rollout under the standard protocol, so a suite total is a binomial proportion over the $N$ perturbed tasks of that suite: $N=2{,}402$ for Spatial, $2{,}518$ for Object, $2{,}591$ for Goal and $2{,}519$ for Long. Table~\ref{tab:wilson} reports approximate Wilson 95\% score intervals for every model and suite total in the LIBERO-Plus comparison, with success counts reconstructed from the published percentages and the suite sizes, and Table~\ref{tab:margins} reports the margin of CoTinyVLA over the strongest baseline in each suite with an interval for that difference.

Interval half-widths lie between 1.1 and 2.0 points throughout, narrowing towards the extremes of the scale where the binomial variance is smallest. A suite total is therefore resolved to roughly two points: differences of five points and above separate cleanly, while differences of one to two points are on the order of the sampling variation.

On Spatial, the interval for CoTinyVLA is $[89.6, 91.9]$, which lies entirely above the intervals of all eleven baselines, including OpenVLA-OFT+ $[84.7, 87.4]$ and RIPT-VLA $[84.4, 87.1]$. The same holds on Object, $[85.9, 88.5]$ against a best baseline interval of $[83.0, 85.9]$, and on Goal, $[85.2, 87.8]$ against $[68.9, 72.4]$. On Long, $[79.1, 82.2]$ lies above ten of the eleven baseline intervals and its lower endpoint sits above the point estimate of the eleventh, OpenVLA-OFT+ (77.7\%). A 0.9B model therefore separates from systems eight times its size on every suite, and the four difference intervals in Table~\ref{tab:margins} all exclude zero.

The per-perturbation results admit the same analysis. Each of the seven axes contributes roughly $N/7 \approx 340$ tasks to a suite. For the largest axis-level margin, Robot Initial States on Goal, 73.6\% against 39.9\%, an independent-normal approximation gives a difference of 33.7 points with a 95\% interval of $[+27.3, +40.1]$; the Language Instructions margin on the same suite, 28.6 points over the strongest 7B baseline, has the interval $[+22.6, +34.6]$. On Spatial the corresponding Robot Initial States margin is 5.7 points with the interval $[-1.7, +13.0]$, so the suite totals separate more sharply than that single axis does. The 12.3-point drop on the Language Instructions axis when paraphrase augmentation is removed is comparable in size to the axis-level interval widths.

Success counts are reconstructed from percentages rounded to one decimal place, which introduces an integer-level ambiguity in the counts that is small relative to the interval widths. The margin intervals use an independent-normal approximation for the difference of two proportions; a paired comparison would require the per-task outcomes of both models, which are unavailable for the baselines.

\begin{table*}[!t]
\centering
\fontsize{9}{10.5}\selectfont
\setlength{\tabcolsep}{5pt}
\renewcommand{\arraystretch}{1.02}
\begin{tabular}{@{}l r r r c@{}}
\toprule
\textbf{Model} & \textbf{$N$} & \textbf{Success} & \textbf{SR (\%)} & \textbf{Wilson 95\% CI} \\
\midrule
\multicolumn{5}{@{}l}{\emph{Spatial}} \\
\midrule
\textbf{CoTinyVLA} & 2402 & 2181 & 90.8 & [89.6, 91.9] \\
OpenVLA-OFT+ & 2402 & 2068 & 86.1 & [84.7, 87.4] \\
RIPT-VLA & 2402 & 2061 & 85.8 & [84.4, 87.1] \\
OpenVLA-OFT & 2402 & 2018 & 84.0 & [82.5, 85.4] \\
OpenVLA-OFT$_m$ & 2402 & 1811 & 75.4 & [73.6, 77.1] \\
$\pi_0$-Fast & 2402 & 1787 & 74.4 & [72.6, 76.1] \\
OpenVLA-OFT$_w$ & 2402 & 1501 & 62.5 & [60.5, 64.4] \\
$\pi_0$ & 2402 & 1458 & 60.7 & [58.7, 62.6] \\
UniVLA & 2402 & 1333 & 55.5 & [53.5, 57.5] \\
NORA & 2402 & 1143 & 47.6 & [45.6, 49.6] \\
WorldVLA & 2402 & 781 & 32.5 & [30.7, 34.4] \\
OpenVLA & 2402 & 466 & 19.4 & [17.9, 21.0] \\
\midrule
\multicolumn{5}{@{}l}{\emph{Object}} \\
\midrule
\textbf{CoTinyVLA} & 2518 & 2198 & 87.3 & [85.9, 88.5] \\
OpenVLA-OFT+ & 2518 & 2128 & 84.5 & [83.0, 85.9] \\
OpenVLA-OFT$_m$ & 2518 & 1941 & 77.1 & [75.4, 78.7] \\
$\pi_0$-Fast & 2518 & 1831 & 72.7 & [70.9, 74.4] \\
OpenVLA-OFT & 2518 & 1674 & 66.5 & [64.6, 68.3] \\
RIPT-VLA & 2518 & 1619 & 64.3 & [62.4, 66.1] \\
$\pi_0$ & 2518 & 1546 & 61.4 & [59.5, 63.3] \\
OpenVLA-OFT$_w$ & 2518 & 1410 & 56.0 & [54.1, 57.9] \\
UniVLA & 2518 & 924 & 36.7 & [34.8, 38.6] \\
NORA & 2518 & 866 & 34.4 & [32.6, 36.3] \\
WorldVLA & 2518 & 720 & 28.6 & [26.9, 30.4] \\
OpenVLA & 2518 & 353 & 14.0 & [12.7, 15.4] \\
\bottomrule
\end{tabular}
\hfill
\begin{tabular}{@{}l r r r c@{}}
\toprule
\textbf{Model} & \textbf{$N$} & \textbf{Success} & \textbf{SR (\%)} & \textbf{Wilson 95\% CI} \\
\midrule
\multicolumn{5}{@{}l}{\emph{Goal}} \\
\midrule
\textbf{CoTinyVLA} & 2591 & 2243 & 86.6 & [85.2, 87.8] \\
OpenVLA-OFT+ & 2591 & 1832 & 70.7 & [68.9, 72.4] \\
OpenVLA-OFT & 2591 & 1632 & 63.0 & [61.1, 64.8] \\
RIPT-VLA & 2591 & 1503 & 58.0 & [56.1, 59.9] \\
$\pi_0$-Fast & 2591 & 1490 & 57.5 & [55.6, 59.4] \\
OpenVLA-OFT$_m$ & 2591 & 1456 & 56.2 & [54.3, 58.1] \\
OpenVLA-OFT$_w$ & 2591 & 1381 & 53.3 & [51.4, 55.2] \\
$\pi_0$ & 2591 & 1163 & 44.9 & [43.0, 46.8] \\
UniVLA & 2591 & 1055 & 40.7 & [38.8, 42.6] \\
NORA & 2591 & 1005 & 38.8 & [36.9, 40.7] \\
WorldVLA & 2591 & 824 & 31.8 & [30.0, 33.6] \\
OpenVLA & 2591 & 391 & 15.1 & [13.8, 16.5] \\
\midrule
\multicolumn{5}{@{}l}{\emph{Long}} \\
\midrule
\textbf{CoTinyVLA} & 2519 & 2033 & 80.7 & [79.1, 82.2] \\
OpenVLA-OFT+ & 2519 & 1957 & 77.7 & [76.0, 79.3] \\
RIPT-VLA & 2519 & 1700 & 67.5 & [65.6, 69.3] \\
OpenVLA-OFT & 2519 & 1673 & 66.4 & [64.5, 68.2] \\
OpenVLA-OFT$_m$ & 2519 & 1610 & 63.9 & [62.0, 65.8] \\
OpenVLA-OFT$_w$ & 2519 & 1315 & 52.2 & [50.3, 54.1] \\
$\pi_0$ & 2519 & 1219 & 48.4 & [46.4, 50.3] \\
$\pi_0$-Fast & 2519 & 1093 & 43.4 & [41.5, 45.3] \\
UniVLA & 2519 & 1005 & 39.9 & [38.0, 41.8] \\
NORA & 2519 & 914 & 36.3 & [34.4, 38.2] \\
OpenVLA & 2519 & 360 & 14.3 & [13.0, 15.7] \\
WorldVLA & 2519 & 207 & 8.2 & [7.2, 9.4] \\
\bottomrule
\end{tabular}
\caption{\textbf{Wilson 95\% confidence intervals for LIBERO-Plus suite totals.} Success counts are reconstructed from the reported suite-level percentages and the number of perturbed tasks per suite, and intervals are Wilson score intervals for a single binomial proportion. $N$ is the number of perturbed tasks in the suite, evaluated with one rollout each under the standard LIBERO-Plus protocol. Interval half-widths are between 1.1 and 2.0 points across all suites and models, which sets the resolution at which suite totals can be compared.}
\label{tab:wilson}
\end{table*}

\begin{table}[!t]
\centering
\fontsize{9}{10.5}\selectfont
\setlength{\tabcolsep}{2.6pt}
\renewcommand{\arraystretch}{1.05}
\begin{tabular}{@{}l r l r c@{}}
\toprule
\textbf{Suite} & \textbf{Ours} & \textbf{Baseline} & \textbf{SR} & \textbf{Margin (95\% CI)} \\
\midrule
Spatial & 90.8 & OpenVLA-OFT+ & 86.1 & $+$4.70 [$+$2.90, $+$6.50] \\
Object & 87.3 & OpenVLA-OFT+ & 84.5 & $+$2.79 [$+$0.87, $+$4.71] \\
Goal & 86.6 & OpenVLA-OFT+ & 70.7 & $+$15.87 [$+$13.68, $+$18.06] \\
Long & 80.7 & OpenVLA-OFT+ & 77.7 & $+$3.01 [$+$0.77, $+$5.25] \\
\bottomrule
\end{tabular}
\caption{\textbf{Suite-level margin of CoTinyVLA over the strongest reported baseline.} The margin interval uses an independent-normal approximation for the difference of two proportions. All four intervals exclude zero.}
\label{tab:margins}
\end{table}

\section{Training Setup}
\label{app:training}

All models in this paper are trained on a single node with eight NVIDIA H100 GPUs. One full training run over the union of the four LIBERO-Plus suites, approximately 2.4\,M action-chunk samples for two epochs, takes about 40 hours of wall time. The ablation study below requires nine such runs, so we train that sweep, including its reference model, for a single epoch; this halves the cost per variant and keeps the study within our compute budget. All nine runs share the same recipe, hyperparameters and schedule and differ only in the dimension under study, so the absolute success rates of Table~\ref{tab:ablation} sit below the two-epoch model of Table~\ref{tab:lp-spatial} while the differences between rows remain matched. Teacher label generation is a separate one-off pass and is not included in this figure. The inference measurements reported later in this supplement are taken on a single NVIDIA L40S, which is the deployment-oriented setting the memory and latency figures are meant to characterise.

\section{Ablation Study}
\label{app:ablation}

Table~\ref{tab:ablation} isolates the contribution of each design choice in CoTinyVLA on LIBERO-Plus Spatial. Each variant differs from the reference in exactly one dimension and is otherwise trained under the identical recipe. To keep the nine-run sweep within budget it is trained for one epoch instead of two, as described under Training Setup, so the reference reaches 87.6\% here against 90.8\% for the two-epoch model of Table~\ref{tab:lp-spatial}; all comparisons below are against the shared one-epoch reference.

First, the visual input configuration (group~(c)) produces the largest total-score reductions in the table. Keeping only the wrist view costs 9.5 points and keeping only the third-person view costs 3.5 points, so the two views are complementary, not redundant, with the third-person view providing broader scene context. Replacing the eight-frame history with a single frame while keeping both views costs 4.7 points, while halving the per-view history from eight to four frames costs only 0.9 points. Temporal context therefore contributes meaningfully, but with diminishing returns past roughly 0.2\,s of history. Across all four input ablations, the Robot Initial States column degrades fastest (down to 51.1 in the single-frame setting), so the temporal input contributes to robustness under kinematic perturbation.

Second, hierarchical CoT distillation (group~(a)) affects several perturbation dimensions. Removing it entirely costs 3.4 points overall, with the largest per-dimension drops on Robot Initial States ($-10.8$) and Objects Layout ($-5.7$), the two lowest-scoring dimensions of LIBERO-Plus Spatial. The two halves of the hierarchy are not symmetric: removing the chunk-level \textsc{Think} (keeping only the episode-level \textsc{Plan}) is more harmful ($-2.2$) than removing the \textsc{Plan} (keeping only \textsc{Think}, $-1.3$). This is consistent with \textsc{Think} supplying supervision at the same temporal resolution as the action chunk, through its explicit \textsc{Phase}, \textsc{Gripper} and \textsc{Next} slots.

Third, paraphrase-based instruction augmentation (group~(b)) primarily affects the Language Instructions axis, consistent with the design rationale in \S\ref{sec:method-aug}. The overall drop without augmentation is moderate ($-2.0$), but the Language Instructions column drops sharply from 86.9 to 74.6 ($-12.3$), while the six non-linguistic perturbation columns are essentially unchanged. The augmentation therefore acts as a localised linguistic-robustness signal, not a general regulariser.

\begin{table*}[t]
\centering
\fontsize{9}{10.5}\selectfont
\setlength{\tabcolsep}{3.5pt}
\renewcommand{\arraystretch}{1.05}
\begin{tabular}{l r r r r r r r r r}
\toprule
\textbf{Variant} & \textbf{Camera} & \textbf{Robot} & \textbf{Lang.} & \textbf{Light} & \textbf{Backgr.} & \textbf{Noise} & \textbf{Layout} & \textbf{Total} & $\Delta$\textbf{Total} \\
\midrule
\textbf{Full CoTinyVLA}                & \textbf{90.4} & \textbf{67.1} & 86.9 & 99.0 & 99.2 & 90.3 & \textbf{85.2} & \textbf{87.6} & -- \\
\midrule
\multicolumn{10}{l}{\emph{(a) Hierarchical chain-of-thought distillation}} \\
No CoT distillation                    & 89.4 & 56.3 & 83.1 & 99.3 & 99.6 & 89.2 & 79.5 & 84.2 & $-3.4$ \\
No \textsc{Plan}, only \textsc{Think}  & 90.7 & 62.6 & 85.4 & 99.3 & 99.6 & 90.0 & 82.6 & 86.3 & $-1.3$ \\
No \textsc{Think}, only \textsc{Plan}  & 89.9 & 58.6 & 87.4 & 99.3 & 99.2 & 89.2 & 80.3 & 85.4 & $-2.2$ \\
\midrule
\multicolumn{10}{l}{\emph{(b) Paraphrase-based instruction augmentation}} \\
No paraphrase augmentation             & 91.0 & 66.3 & 74.6 & 99.3 & 99.6 & 90.9 & 84.4 & 85.6 & $-2.0$ \\
\midrule
\multicolumn{10}{l}{\emph{(c) Dual-view temporal visual input}} \\
Third-person view only                 & 84.6 & 56.9 & 85.6 & 99.3 & 99.6 & 88.9 & 80.5 & 84.1 & $-3.5$ \\
Wrist view only                        & 78.2 & 53.1 & 80.8 & 95.2 & 96.1 & 80.9 & 70.4 & 78.1 & $-9.5$ \\
Single frame, dual-view                & 85.1 & 51.1 & 87.4 & 99.3 & 99.6 & 87.2 & 77.4 & 82.9 & $-4.7$ \\
4 frames per view (instead of 8)       & 91.2 & 62.0 & 86.7 & 99.3 & 99.6 & 90.6 & 83.1 & 86.7 & $-0.9$ \\
\bottomrule
\end{tabular}
\caption{\textbf{Ablation study on LIBERO-Plus Spatial} (success rate, \%). Each variant modifies a single design choice from the reference model and is otherwise trained and evaluated under the identical recipe. The sweep is trained for one epoch instead of the two epochs used for the model in Table~\ref{tab:lp-spatial}, which keeps nine training runs within budget; the reference row therefore sits below Table~\ref{tab:lp-spatial} in absolute terms while all rows remain matched to each other. Bold marks the best entry per column.}
\label{tab:ablation}
\end{table*}

\section{Closed-Loop Inference}
\label{app:freerun}

Table~\ref{tab:freerun} summarises 80 closed-loop episodes, 20 on each of the four standard LIBERO suites, in which the model generates one \textsc{Plan} per episode and one \textsc{Think} span per action chunk under greedy decoding with a 120-token budget. A policy inference step comprises image preprocessing and tokenisation, autoregressive generation of the reasoning span, the vision-language forward pass, the proprioception projector, the action head, and the generation KV cache; every figure below covers it end to end.

\paragraph{Output-format reliability.} All 1{,}552 generated spans were parseable and none reached the token cap. A run of 1{,}552 clean outputs places a one-sided 95\% upper bound of about 0.19\% on the malformed rate, so the structure imposed by the distillation objective at training time is reproduced at inference. A malformed span changes the representation the action head consumes, so format validity is an interface requirement, not a cosmetic property, and a schema closed at the \textsc{Gripper} slot and bounded by the \textsc{Plan} at the \textsc{Phase} slot is enough to keep a 0.9B model inside it.

\paragraph{Closed-loop success.} The policy succeeds in 79 of 80 episodes, or 98.8\% with a Wilson interval of $[93.3, 99.8]$: 20 of 20 on Spatial, Object and Goal, and 19 of 20 on the long-horizon suite, where the single failure exhausts the step budget.

\paragraph{Latency decomposition.} The cost profile is strongly asymmetric (Table~\ref{tab:latencydecomp}). A steady-state chunk takes 1.37\,s at the median and 1.49\,s at the 95th percentile, and 76\% of that is autoregressive generation of 26 reasoning tokens. Everything else combined, including the vision-language forward pass over 16 images, is 276\,ms. The episode-initial chunk, which also generates the \textsc{Plan}, takes 2.76\,s with 88\% in generation. For a model of this size the dominant cost at inference is therefore not the visual stream and not the backbone forward pass, but the serial decoding of a few dozen tokens. The current implementation is synchronous, so a chunk of eight actions is emitted every 1.37\,s. Reducing that number is therefore a decoding problem, not an architecture problem, and admits standard remedies that leave the trained weights unchanged; the next section quantifies one of them.

\paragraph{Memory usage.} Peak allocated memory is 2.25\,GiB at the median and 2.27\,GiB at the maximum across all four suites, only 0.05\,GiB above the value measured without generation, so the generation KV cache for a span of this length is a negligible addition to a model whose footprint is dominated by its weights. Peak \emph{reserved} memory, which the PyTorch caching allocator holds from the driver and exceeds what the model requires, stays within 2.30\,GiB.

\begin{table*}[!t]
\centering
\fontsize{9}{10.5}\selectfont
\setlength{\tabcolsep}{5pt}
\renewcommand{\arraystretch}{1.05}
\begin{tabular}{@{}l r r r r r r r r r r r@{}}
\toprule
& & & & \multicolumn{2}{c}{\textbf{Format}} & \multicolumn{3}{c}{\textbf{Chunk latency (ms)}} & \textbf{Generated} & \multicolumn{2}{c}{\textbf{GPU memory (GiB)}} \\
\cmidrule(lr){5-6}\cmidrule(lr){7-9}\cmidrule(lr){11-12}
\textbf{Suite} & \textbf{Success} & \textbf{\%} & \textbf{Chunks} & \textbf{Malf.} & \textbf{Trunc.} & \textbf{initial} & \textbf{steady} & \textbf{steady p95} & \textbf{tokens} & \textbf{Allocated} & \textbf{Reserved} \\
\midrule
Spatial & 20/20 & 100 & 264 & 0 & 0 & 2746 & 1366 & 1484 & 25.9 & 2.24 & 2.28 \\
Object & 20/20 & 100 & 337 & 0 & 0 & 2648 & 1362 & 1557 & 26.4 & 2.24 & 2.28 \\
Goal & 20/20 & 100 & 299 & 0 & 0 & 2543 & 1364 & 1500 & 25.5 & 2.24 & 2.28 \\
Long & 19/20 & 95 & 652 & 0 & 0 & 3789 & 1388 & 1573 & 26.9 & 2.26 & 2.29 \\
\midrule
\textbf{All four suites} & \textbf{79/80} & \textbf{98.75} & \textbf{1552} & \textbf{0} & \textbf{0} & 2763 & 1370 & 1489 & 26.0 & 2.25 & 2.28 \\
\bottomrule
\end{tabular}
\caption{\textbf{Closed-loop inference on the four standard LIBERO suites.} The model generates one \textsc{Plan} per episode and one \textsc{Think} span per action chunk; no ground-truth intermediate tokens are supplied. 20 episodes per suite, greedy decoding, 120-token reasoning budget, a uniform 600-step rollout cap, FlashAttention-2, single seed, NVIDIA L40S. \textbf{Malf.}\ and \textbf{Trunc.}\ are the percentages of action chunks whose reasoning span failed to parse or reached the token cap. \emph{Initial} chunks generate a fresh \textsc{Plan} and \textsc{Think}; \emph{steady} chunks reuse the episode \textsc{Plan} and generate only \textsc{Think}. Latency and memory cover the complete policy inference step, including tokenisation, autoregressive generation, the vision-language forward pass, the proprioception projector, the action head, and the generation KV cache; three rollouts share each GPU, so latencies are measured under that fixed concurrency.}
\label{tab:freerun}
\end{table*}

\begin{table}[!t]
\centering
\fontsize{9}{10.5}\selectfont
\setlength{\tabcolsep}{3pt}
\renewcommand{\arraystretch}{1.05}
\begin{tabular}{@{}l r r r r@{}}
\toprule
& \multicolumn{2}{c}{\textbf{Initial}} & \multicolumn{2}{c}{\textbf{Steady state}} \\
\cmidrule(lr){2-3}\cmidrule(lr){4-5}
\textbf{Stage} & \textbf{ms} & \textbf{\%} & \textbf{ms} & \textbf{\%} \\
\midrule
Preprocessing and tokenisation & 29.7 & 1.1 & 24.7 & 1.8 \\
Reasoning generation & 2437.1 & 88.2 & 1047.4 & 76.4 \\
Vision-language forward & 252.7 & 9.1 & 249.9 & 18.2 \\
Proprio.\ projector & 0.6 & 0.0 & 0.5 & 0.0 \\
Action head & 0.6 & 0.0 & 0.5 & 0.0 \\
\midrule
\textbf{Total policy inference step} & \textbf{2763} & \textbf{100.0} & \textbf{1370} & \textbf{100.0} \\
Generated tokens & 69 & -- & 26 & -- \\
\bottomrule
\end{tabular}
\caption{\textbf{Where the time goes in a closed-loop policy inference step.} Median over the 80 episode-initial and 1{,}472 steady-state chunks of Table~\ref{tab:freerun}. An initial chunk generates both the episode \textsc{Plan} and the first \textsc{Think}; a steady-state chunk reuses the cached \textsc{Plan} and generates only \textsc{Think}. Autoregressive generation accounts for 88\% of the initial step and 76\% of a steady-state step, and every other stage together contributes under 280\,ms.}
\label{tab:latencydecomp}
\end{table}

\section{The Cost of the Reasoning Span}
\label{app:cotcost}

Generation dominates inference cost, and two variables control it: how often the \textsc{Plan} is regenerated and how many tokens the policy may emit. We vary both on LIBERO Spatial under closed-loop inference (Table~\ref{tab:cotcost}).

Caching the episode-level \textsc{Plan} reduces median steady-state latency from 2.65\,s to 1.37\,s and the mean generated length from 70.3 to 25.9 tokens, a 48.5\% latency reduction and a 63.2\% token reduction, with success unchanged at 100\%. The episode-level \textsc{Plan} is therefore not merely a convenient way to organise the supervision: because the task-level intent is by construction stable within an episode, emitting it once and conditioning every subsequent chunk on the cached text removes almost half of the inference cost with success unchanged at 100\%.

The token budget affects performance only once spans are truncated. At a 48-token cap, 28.0\% of chunks are truncated and success falls to 90\%; at 64 tokens truncation drops to 6.0\% and success to 95\%; at 96 and 120 tokens no chunk is truncated and the two configurations are indistinguishable in success, latency and generated length. Steady-state latency is flat at 1.37\,s across all four budgets, because the policy stops when it emits the closing tag, before it reaches the cap, so a smaller cap buys nothing and only risks truncation. Ninety-six tokens is therefore the smallest budget that leaves the reasoning span intact. The 48-token result also confirms that truncation is detected by the parser: when the budget is too small the malformed rate rises to 28.0\%.

\begin{table}[!t]
\centering
\fontsize{9}{10.5}\selectfont
\setlength{\tabcolsep}{2.2pt}
\renewcommand{\arraystretch}{1.05}
\begin{tabular}{@{}l r r r r r r@{}}
\toprule
& & \textbf{SR} & \textbf{Trunc.} & \multicolumn{2}{c}{\textbf{Latency (ms)}} & \textbf{Gen.} \\
\cmidrule(lr){5-6}
\textbf{Configuration} & \textbf{Cap} & \textbf{(\%)} & \textbf{(\%)} & \textbf{init.} & \textbf{steady} & \textbf{tok.} \\
\midrule
\multicolumn{7}{@{}l}{\emph{\textsc{Plan} caching (120-token cap)}} \\
\midrule
\textbf{Cached \textsc{Plan}} & 120 & 100 & 0.0 & 2746 & 1366 & 25.9 \\
\textsc{Plan} per chunk & 120 & 100 & 0.0 & 2684 & 2653 & 70.3 \\
\midrule
\multicolumn{7}{@{}l}{\emph{Reasoning-token budget}} \\
\midrule
Budget 48 & 48 & 90 & 28.0 & 2042 & 1376 & 31.1 \\
Budget 64 & 64 & 95 & 6.0 & 2592 & 1365 & 25.9 \\
Budget 96 & 96 & 100 & 0.0 & 2755 & 1367 & 25.9 \\
\textbf{Budget 120} & 120 & 100 & 0.0 & 2746 & 1366 & 25.9 \\
\bottomrule
\end{tabular}
\caption{\textbf{Cost of the reasoning span on LIBERO Spatial}, 20 episodes per configuration under closed-loop inference. \textbf{Cap} is the maximum number of reasoning tokens the policy may emit and \textbf{Trunc.}\ the percentage of chunks that reach it. Caching the episode \textsc{Plan} halves steady-state latency and cuts generated tokens by 63\% at identical success; the token budget can be reduced to 96 without truncation, while 48 and 64 truncate and lose success.}
\label{tab:cotcost}
\end{table}

\section{Test-Time Interventions on the Trained Policy}
\label{app:interventions}

Where the ablation study retrains the model with a component removed, the interventions in Table~\ref{tab:interventions} hold the trained weights fixed and perturb the inputs at test time, on the same episodes as the closed-loop reference, so every comparison is paired and can be tested exactly. Because the reference succeeds on all 40 Spatial and Goal episodes, any drop is attributable to the intervention. For the interventions that rewrite the reasoning span, we also report the median $L_2$ distance between the action chunk emitted with and without the intervention, which measures the effect in the continuous output space and not only at the level of task outcome.

\paragraph{\textsc{Plan} content is load-bearing.} Replacing the episode \textsc{Plan} with an empty span reduces pooled success from 100\% to 60\%, and replacing it with a \textsc{Plan} for a different task reduces it to 55\% ($p<0.001$ in both cases). The effect is larger on Goal (40\% and 35\%) than on Spatial (80\% and 75\%), which is what the two suites ask for: a Goal task is a sequence of subgoals whose order the \textsc{Plan} encodes, whereas a Spatial task is closer to a single reach-and-place whose intent is recoverable from the instruction alone. The intervention moves the emitted actions as well as the outcomes, by a median $L_2$ of 0.73 to 0.85 at a cosine similarity of 0.96 to 0.98. The episode-level component is therefore not a readable annotation attached to a policy that would behave the same without it: its content enters the representation the action head consumes. This is the role the schema of \S\ref{sec:method-cot} assigns to it, and it is also what makes the \textsc{Plan} cache worthwhile, since the episode-level intent it encodes is stable across the chunks that reuse it.

\paragraph{The reasoning span is on the execution path.} Removing the reasoning span altogether takes success to 0\% on both suites while cutting steady-state latency from 1.37\,s to 0.29\,s. Action prediction therefore depends on a reasoning-conditioned readout. Removing the span also changes the position from which the action head reads, so this result concerns the readout pathway.

\paragraph{Temporal variation within the history.} Replacing the eight-frame window with the most recent frame repeated eight times, which leaves the token count, the memory footprint and the latency unchanged, reduces pooled success from 100\% to 60\% ($p<0.001$); the gain from the temporal input is therefore not attributable to the number of image tokens. Reversing the order of the history is more selective, taking Spatial from 100\% to 60\% while leaving Goal at 85\%. The Spatial tasks turn on where objects and the gripper are relative to one another, so the direction of recent motion is informative; the Goal tasks turn on which of several arrangements is being assembled, for which the set of observed configurations matters more than its order. Both interventions also destabilise the reasoning trace, raising the \textsc{Phase} regression rate from 0.8\% to 15.2\% on Spatial, so the degradation is visible in the model's own output before it is visible in the outcome.

\paragraph{Relative run-time importance of the two views.} Occluding the wrist view takes both suites to 0\%, while occluding the third-person view leaves Spatial at 100\% and Goal at 95\%. Gripper--proprioception agreement falls from 90.9\% to 31.9\% on Spatial under wrist occlusion, so the model loses track of its own end effector before it loses the task. The retraining ablation of \S\ref{app:ablation} orders the views the other way, with a wrist-only model losing 9.5 points against 3.5 for a third-person-only model, and the two measurements answer different questions: a model trained on both streams comes to depend on the closer view for fine positioning, while a model that must be trained on one stream alone does better with the view that carries global scene context.

\paragraph{Robustness to the evaluated instruction variants.} Replacing each instruction with a paraphrase drawn from the augmentation set, and prefixing a politeness marker, both leave success at 100\% on both suites. Together with the retraining ablation, this indicates robustness to paraphrases of the kind the augmentation supplies.

\begin{table*}[!t]
\centering
\fontsize{9}{10.5}\selectfont
\setlength{\tabcolsep}{6pt}
\renewcommand{\arraystretch}{1.05}
\begin{tabular}{@{}l r r r r r r@{}}
\toprule
& \multicolumn{3}{c}{\textbf{Success rate (\%)}} & & & \textbf{Action $L_2$} \\
\cmidrule(lr){2-4}
\textbf{Intervention} & \textbf{Spatial} & \textbf{Goal} & \textbf{pooled} & \textbf{$\Delta$} & \textbf{$p$} & \textbf{Spat.\ / Goal} \\
\midrule
\multicolumn{7}{@{}l}{\emph{Reference}} \\
\midrule
Unmodified policy & 100 & 100 & 100.0 & -- & -- & -- \\
\midrule
\multicolumn{7}{@{}l}{\emph{Reasoning span}} \\
\midrule
\textsc{Plan} replaced by an empty span & 80 & 40 & 60.0 & $-$40 & $<$0.001 & 0.73 / 0.85 \\
\textsc{Plan} replaced by a contradictory one & 75 & 35 & 55.0 & $-$45 & $<$0.001 & 0.73 / 0.79 \\
Reasoning span removed & 0 & 0 & 0.0 & $-$100 & $<$0.001 & -- / -- \\
\midrule
\multicolumn{7}{@{}l}{\emph{Observation history}} \\
\midrule
Most recent frame repeated $16\times$ & 75 & 45 & 60.0 & $-$40 & $<$0.001 & -- / -- \\
History order reversed & 60 & 85 & 72.5 & $-$27.5 & $<$0.001 & -- / -- \\
\midrule
\multicolumn{7}{@{}l}{\emph{Camera views}} \\
\midrule
Wrist view occluded & 0 & 0 & 0.0 & $-$100 & $<$0.001 & -- / -- \\
Third-person view occluded & 100 & 95 & 97.5 & $-$2.5 & 1.00 & -- / -- \\
\midrule
\multicolumn{7}{@{}l}{\emph{Instruction wording}} \\
\midrule
Training-set paraphrase & 100 & 100 & 100.0 & $\pm$0 & 1.00 & -- / -- \\
Politeness prefix added & 100 & 100 & 100.0 & $\pm$0 & 1.00 & -- / -- \\
\bottomrule
\end{tabular}
\caption{\textbf{Test-time interventions on the trained policy.} Each intervention is applied at inference only, on the same 20 episodes per suite as the unmodified reference, so every comparison is paired; $p$ is an exact McNemar test on the 40 pooled episode outcomes. \textbf{Action $L_2$} is the median distance between the action chunk the policy emits with and without the intervention, reported for the interventions that alter the reasoning span. Occlusion replaces one camera stream with black frames, which is outside the training distribution, so those rows indicate which stream the trained policy relies on at run time, not the accuracy of a model trained without that view.}
\label{tab:interventions}
\end{table*}

\section{Internal Consistency of the Generated Reasoning}
\label{app:reasoncons}

Because the reasoning span is emitted in a fixed schema, two of its slots can be checked automatically against quantities the simulator already provides (Table~\ref{tab:reasoncons}).

The \textsc{Gripper} slot agrees with the proprioception reading in 82.7\% of chunks overall, but the aggregate is dominated by a boundary case. Agreement is 97.6\% when the gripper is open and 82.4\% when it is closed, and falls to 29.7\% in the transient \textsc{PARTIALLY\_CLOSED} state that occurs while the fingers are still moving; excluding that state, agreement is 94.3\%. Most disagreement therefore occurs where a discrete label is least well defined, which reflects the ambiguity of discretising a continuous joint gap.

The \textsc{Phase} slot tracks progress through the episode. A \emph{regression} occurs when the emitted step index moves backwards relative to the previous chunk, that is, when the model revises its belief about how far through the task it is. Regressions are rare under normal operation, 51 of 1{,}552 chunks, and they respond sharply to the interventions that cost success: repeating the latest frame raises the Spatial rate from 0.8\% to 15.2\%, reversing the history to 15.0\% and emptying the \textsc{Plan} to 13.0\%, while occluding the third-person view, which leaves success at 100\%, leaves the rate at 1.2\%. The rate is therefore observable at run time without access to the reward, and it moves with the conditions the policy depends on. The closed-loop reference times out on a single episode, so the association with task failure is not examined here.

\begin{table}[!t]
\centering
\fontsize{9}{10.5}\selectfont
\setlength{\tabcolsep}{5pt}
\renewcommand{\arraystretch}{1.05}
\begin{tabular}{@{}l r r r@{}}
\toprule
\multicolumn{4}{@{}l}{\emph{(a) \textsc{Gripper} slot against the proprioception reading}} \\
\midrule
\textbf{Proprioception state} & \textbf{Chunks} & \textbf{Agree} & \textbf{\%} \\
\midrule
\textsc{OPEN} & 994 & 970 & 97.6 \\
\textsc{CLOSED} & 279 & 230 & 82.4 \\
\textsc{PARTIALLY\_CLOSED} & 279 & 83 & 29.7 \\
\midrule
All states & 1552 & 1283 & 82.7 \\
Excluding \textsc{PARTIALLY\_CLOSED} & 1273 & 1200 & 94.3 \\
\bottomrule
\end{tabular}

\vspace{0.6em}
\begin{tabular}{@{}l r r@{}}
\toprule
\multicolumn{3}{@{}l}{\emph{(b) \textsc{Phase} regressions on Spatial}} \\
\midrule
\textbf{Condition} & \textbf{Chunks} & \textbf{Regr.\ (\%)} \\
\midrule
Unmodified policy & 264 & 0.82 \\
Most recent frame repeated & 716 & 15.23 \\
History order reversed & 812 & 15.03 \\
\textsc{Plan} replaced by an empty span & 627 & 13.01 \\
Third-person view occluded & 271 & 1.20 \\
\bottomrule
\end{tabular}
\caption{\textbf{Internal consistency of the generated reasoning.} (a)~compares the emitted \textsc{Gripper} slot with the proprioception reading at the same step, over the 1{,}552 closed-loop chunks of Table~\ref{tab:freerun}; \textsc{PARTIALLY\_CLOSED} is the transient state between the other two. (b)~counts \textsc{Phase} regressions, where the emitted step index moves backwards relative to the previous chunk of the same episode, for the unmodified policy and for four of the test-time interventions of Table~\ref{tab:interventions} on the Spatial suite.}
\label{tab:reasoncons}
\end{table}

\section{Cost of the Visual Input Stream}
\label{app:acccost}

The closed-loop measurements above cover the complete policy inference step and are dominated by decoding, so they are the relevant quantity for deployment but do not isolate the contribution of the visual input. To separate the visual stream we profile the forward pass alone, with the reasoning span supplied as input instead of generated, so that the only variable is the number and arrangement of images. Table~\ref{tab:efficiency} reports this profile for five input configurations at batch size~1 in \texttt{bfloat16} on a single NVIDIA L40S, with 10 warm-up iterations and 50 timed repetitions over 32 held-out inputs. The 249.2\,ms of the full configuration sits inside the 275\,ms of vision-language forward pass and preprocessing reported in Table~\ref{tab:latencydecomp}; the remainder of a policy inference step is generation.

Peak memory increases approximately linearly with image count: the resident allocation is 1.61--1.62\,GiB in every configuration and is largely attributable to persistent model allocations, while increasing the input from two images to 16 raises the peak from 1.76\,GiB to 2.20\,GiB, an increment of roughly 32\,MiB per image. Forward time scales at about 6.5\,ms and 57 input tokens per additional image, at a stable rate across the whole range. For a fixed number of images, latency and memory are similar across view allocations: the three eight-image configurations agree to within 1.6\,ms of forward time and 0.01\,GiB of peak memory, which makes the comparison of input structure matched in cost.

\begin{table*}[!t]
\centering
\fontsize{9}{10.5}\selectfont
\setlength{\tabcolsep}{6pt}
\renewcommand{\arraystretch}{1.05}
\begin{tabular}{l r r r r r r r r r}
\toprule
& & \textbf{Input} & \multicolumn{3}{c}{\textbf{Forward pass (ms)}} & \textbf{Amortised} & \textbf{Env.\ steps} & \multicolumn{2}{c}{\textbf{GPU memory (GiB)}} \\
\cmidrule(lr){4-6}\cmidrule(lr){9-10}
\textbf{Input configuration} & \textbf{Images} & \textbf{tokens} & \textbf{median} & \textbf{p05} & \textbf{p95} & \textbf{ms / step} & \textbf{per s} & \textbf{Allocated} & \textbf{Reserved} \\
\midrule
\textbf{Full (8+8)} & 16 & 1083 & 249.2 & 247.0 & 255.6 & 31.2 & 32.1 & 2.20 & 2.22 \\
Half (4+4) & 8 & 627 & 196.6 & 194.6 & 198.9 & 24.6 & 40.7 & 1.95 & 1.97 \\
Third-person only (8+0) & 8 & 614 & 195.8 & 194.8 & 200.1 & 24.5 & 40.9 & 1.94 & 1.97 \\
Wrist only (0+8) & 8 & 622 & 197.4 & 194.3 & 201.6 & 24.7 & 40.5 & 1.95 & 1.97 \\
Single frame (1+1) & 2 & 281 & 158.8 & 157.3 & 160.8 & 19.9 & 50.4 & 1.76 & 1.81 \\
\bottomrule
\end{tabular}
\caption{\textbf{Forward-pass cost of the dual-view temporal input.} Reasoning tokens are supplied as input during this profile and are not generated, so the reported times are the visual and backbone component of a policy inference step, not its total; \S\ref{app:freerun} reports the end-to-end cost with closed-loop generation. Model-level forward profiling of the trained CoTinyVLA checkpoint at batch size~1 in \texttt{bfloat16} on a single NVIDIA L40S (10 warm-up and 50 timed repetitions over 32 held-out inputs). Timing covers the vision encoder, the Qwen3.5-0.8B backbone, the proprioception projector and the action head with the CoT span present in the input, and excludes image decoding from disk and tokenisation. Resident weights account for 1.62\,GiB in every configuration; the remainder of the peak is the activation working set.}
\label{tab:efficiency}
\end{table*}

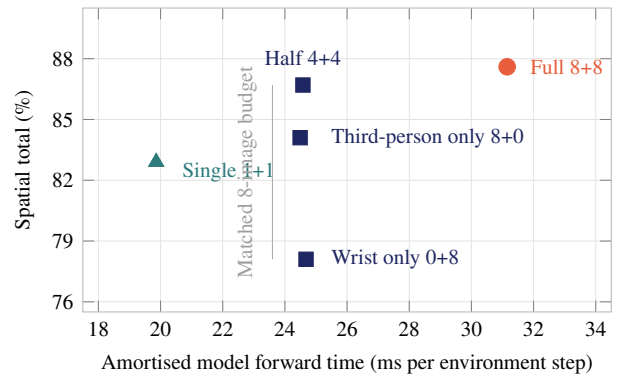
\begin{figure}[!t]
\centering
\begin{tikzpicture}
\begin{axis}[
    width=\columnwidth,
    height=5.6cm,
    xlabel={Amortised model forward time (ms per environment step)},
    ylabel={Spatial total (\%)},
    xlabel style={font=\footnotesize},
    ylabel style={font=\footnotesize},
    tick label style={font=\footnotesize},
    xmin=17.5, xmax=34.5,
    ymin=75.5, ymax=90.5,
    xtick={18,20,22,24,26,28,30,32,34},
    ytick={76,79,82,85,88},
    grid=major,
    grid style={gray!22, line width=0.3pt},
    axis line style={gray!55},
    clip=false,
]
\addplot[only marks, mark=*, mark size=3.1pt, color=coral]
    coordinates {(31.15,87.6)};
\addplot[only marks, mark=square*, mark size=2.7pt, color=navy]
    coordinates {(24.58,86.7) (24.49,84.1) (24.68,78.1)};
\addplot[only marks, mark=triangle*, mark size=3.3pt, color=teal]
    coordinates {(19.86,82.9)};

\node[font=\footnotesize, anchor=west, color=coral] at (axis cs:31.6,87.6) {Full 8+8};
\node[font=\footnotesize, anchor=south, color=navy] at (axis cs:24.58,87.2) {Half 4+4};
\node[font=\footnotesize, anchor=west, color=navy] at (axis cs:25.2,84.1) {Third-person only 8+0};
\node[font=\footnotesize, anchor=west, color=navy] at (axis cs:25.2,78.1) {Wrist only 0+8};
\node[font=\footnotesize, anchor=west, color=teal] at (axis cs:20.4,82.4) {Single 1+1};

\draw[gray!65, line width=0.4pt] (axis cs:23.6,78.1) -- (axis cs:23.6,86.7);
\node[font=\footnotesize, gray!75, rotate=90, anchor=south] at (axis cs:23.4,82.4) {Matched 8-image budget};
\end{axis}
\end{tikzpicture}
\caption{Accuracy against amortised model forward time for the five input configurations of Table~\ref{tab:acccost}. The three eight-image configurations (squares) are matched in cost to within 0.2\,ms per environment step, so their 8.6-point spread on Spatial total isolates the effect of how a fixed visual budget is allocated across cameras and across time.}
\label{fig:acccost}
\end{figure}

Table~\ref{tab:acccost} pairs these measurements with the corresponding entries of the input ablation, and Figure~\ref{fig:acccost} plots accuracy against amortised forward time.

At matched cost, the three eight-image configurations are indistinguishable in latency and memory, yet they reach 86.7\%, 84.1\% and 78.1\% on Spatial depending on whether the budget is split as four frames per camera, spent entirely on the third-person camera, or spent entirely on the wrist camera. View allocation therefore accounts for an 8.6-point spread at matched compute. Splitting the budget across both cameras is the best of the three, so view diversity and temporal depth are complementary, not substitutable: four frames from each camera beat eight frames from either camera alone. The ordering of the two single-camera configurations is also informative: the third-person-only configuration is 6.0 points higher than the wrist-only configuration, matching the account in \S\ref{sec:method-arch} that the third-person camera carries the global scene layout while the wrist camera carries local contact detail, so a model given only the wrist stream must infer the layout it cannot see.

Halving the history from eight frames per view to four reduces forward time from 249.2\,ms to 196.6\,ms and peak memory from 2.20\,GiB to 1.95\,GiB while giving up 0.9 points of Spatial total, a lower-cost operating point for settings where latency or memory is the binding constraint. Reducing further to a single frame per view saves another 37.8\,ms but costs 4.7 points, and the loss is concentrated almost entirely on one axis: Robot Initial States falls from 67.1\% to 51.1\%. Expressed as a rate, the 16.0-point gain in kinematic robustness that the temporal history provides costs 11.3\,ms per environment step. The temporal input is thus not a general-purpose accuracy improvement that happens to be expensive: it is associated with a large, axis-specific gain in robustness to physical-state perturbation for a small measured latency increase, on the axis where compact models are otherwise weakest.

The Language Instructions axis is essentially flat across the four configurations that retain a third-person view, ranging only from 85.6\% to 87.4\%, and it does not track the total at all: the single-frame configuration has the second-lowest total and the highest Language score. Linguistic robustness in CoTinyVLA therefore comes from the paraphrase augmentation and not from the visual stream, as the ablation study also shows. The three design components act on separate axes on the cost side as well as the accuracy side, and across every configuration evaluated here peak allocated memory stays within the 2.3\,GiB envelope measured under closed-loop inference.

\begin{table}[!t]
\centering
\fontsize{9}{10.5}\selectfont
\setlength{\tabcolsep}{3.5pt}
\renewcommand{\arraystretch}{1.05}
\begin{tabular}{@{}l r r r r r r@{}}
\toprule
& \multicolumn{2}{c}{\textbf{Cost}} & \multicolumn{4}{c}{\textbf{LIBERO-Plus Spatial}} \\
\cmidrule(lr){2-3}\cmidrule(lr){4-7}
\textbf{Configuration} & \textbf{ms/step} & \textbf{GiB} & \textbf{Total} & \textbf{$\Delta$} & \textbf{Robot} & \textbf{Lang.} \\
\midrule
\textbf{Full 8+8} & 31.2 & 2.20 & 87.6 & -- & 67.1 & 86.9 \\
Half 4+4 & 24.6 & 1.95 & 86.7 & $-$0.9 & 62.0 & 86.7 \\
Third-person 8+0 & 24.5 & 1.94 & 84.1 & $-$3.5 & 56.9 & 85.6 \\
Wrist 0+8 & 24.7 & 1.95 & 78.1 & $-$9.5 & 53.1 & 80.8 \\
Single 1+1 & 19.9 & 1.76 & 82.9 & $-$4.7 & 51.1 & 87.4 \\
\bottomrule
\end{tabular}
\caption{\textbf{Accuracy--cost profile of the visual input stream.} Amortised per-environment-step forward time and peak allocated memory from Table~\ref{tab:efficiency}, paired with the corresponding LIBERO-Plus Spatial success rates from the input ablation (Table~\ref{tab:ablation}). \textbf{Robot} is the Robot Initial States axis and \textbf{Lang.}\ the Language Instructions axis. The three eight-image configurations are matched in cost to within 0.2\,ms per step and 0.01\,GiB, so the differences among them isolate how a fixed image budget is allocated across cameras and across time. Accuracy is taken from the one-epoch ablation sweep of Table~\ref{tab:ablation}, so the comparison of interest is across configurations, not the absolute level.}
\label{tab:acccost}
\end{table}

\section{Instruction Variants}
\label{app:variants}

The paraphrase-based instruction set covers all 40 base instructions across the four LIBERO-Plus training suites, expanded to 20 paraphrases per base instruction. Paraphrases combine three independent transformations: verb substitution from a controlled vocabulary, object-synonym substitution from a controlled lexicon, and politeness variation drawn from five register templates. A representative example for the Spatial base instruction \emph{``pick up the black bowl next to the plate and place it on the plate''} includes paraphrases such as \emph{``grab the dark bowl beside the plate and set it down on the plate''}, \emph{``could you lift the dark-hued bowl next to the plate and place it onto the plate''}, and \emph{``please relocate the black bowl adjacent to the plate, putting it on top of the plate''}.

Table~\ref{tab:instrcov} summarises the lexical diversity of the paraphrase set per base instruction.

The set contains 800 paraphrases across 40 base instructions, all of them distinct. Averaged over the set, the mean word-set Jaccard similarity to the base instruction is 0.44, the mean character-level Levenshtein similarity ratio is 0.60, and each paraphrase introduces 3.84 tokens that the base instruction does not contain. At this level of surface variation the model rarely sees the same wording twice during training.

The transformations are applied at very different rates. The main verb or command surface differs in 787 of 800 paraphrases, whereas an explicit politeness or register marker appears in 125, roughly one in six. The augmentation is therefore predominantly an intervention on the action-denoting part of the instruction, with register variation as a smaller secondary axis. Verb substitution breaks the recurrence of individual command verbs, which a compact student trained on only 40 surface forms could otherwise use as a shortcut.

The per-instruction breakdown shows a consistent pattern: lexical similarity increases with instruction length. The shortest base instruction, \emph{``turn on the stove''} (four words), has a Jaccard similarity of 0.39 and a Levenshtein ratio of 0.45, while the two longest compositional instructions, which name two objects and two targets each, sit at 0.66--0.68 and 0.71--0.72. Longer instructions carry more object-identifying tokens that a paraphrase must preserve for the command to remain executable in the scene, so the proportion of the instruction that is free to vary shrinks as the referential load grows.

These statistics give the scale of the intervention behind the axis-specific ablation result. Removing the augmentation moves the Language Instructions axis from 86.9\% to 74.6\% while leaving the six non-linguistic axes essentially unchanged, and Table~\ref{tab:acccost} shows that the same axis is insensitive to the visual input configuration. With 800 distinct surface forms spanning verb substitution, synonym substitution and register variation, the augmentation supplies substantial lexical variation around every base instruction.

\begin{table*}[!t]
\centering
\fontsize{9}{10.5}\selectfont
\setlength{\tabcolsep}{5pt}
\renewcommand{\arraystretch}{1.02}
\begin{tabular}{@{}p{0.42\textwidth} r r r r r r r r@{}}
\toprule
\textbf{Base instruction} & \textbf{Var.} & \textbf{Uniq.} & \textbf{Words} & \textbf{Jacc.} & \textbf{Lev.} & \textbf{New} & \textbf{Reg.} & \textbf{Verb} \\
\midrule
turn on the stove & 20 & 20 & 5.05 & 0.39 & 0.45 & 2.55 & 4 & 20 \\
put the bowl on the plate & 20 & 20 & 7.55 & 0.45 & 0.62 & 3.00 & 3 & 20 \\
put the bowl on the stove & 20 & 20 & 7.75 & 0.41 & 0.59 & 3.40 & 3 & 20 \\
put the cream cheese in the bowl & 20 & 20 & 8.45 & 0.50 & 0.68 & 3.00 & 2 & 20 \\
pick up the butter and place it in the basket & 20 & 20 & 8.55 & 0.34 & 0.53 & 3.35 & 5 & 20 \\
pick up the milk and place it in the basket & 20 & 20 & 8.60 & 0.37 & 0.52 & 3.15 & 5 & 20 \\
put the wine bottle on the rack & 20 & 20 & 8.60 & 0.50 & 0.62 & 3.15 & 3 & 20 \\
put the bowl on top of the cabinet & 20 & 20 & 8.80 & 0.40 & 0.59 & 3.60 & 3 & 20 \\
pick up the ketchup and place it in the basket & 20 & 20 & 8.90 & 0.35 & 0.56 & 3.45 & 4 & 20 \\
pick up the salad dressing and place it in the basket & 20 & 20 & 9.15 & 0.38 & 0.59 & 3.10 & 5 & 20 \\
pick up the orange juice and place it in the basket & 20 & 20 & 9.20 & 0.37 & 0.55 & 3.40 & 3 & 20 \\
pick up the chocolate pudding and place it in the basket & 20 & 20 & 9.30 & 0.43 & 0.64 & 2.85 & 2 & 20 \\
open the middle drawer of the cabinet & 20 & 20 & 9.40 & 0.44 & 0.47 & 4.30 & 4 & 19 \\
pick up the tomato sauce and place it in the basket & 20 & 20 & 9.40 & 0.42 & 0.62 & 3.00 & 3 & 20 \\
put both moka pots on the stove & 20 & 20 & 9.45 & 0.46 & 0.60 & 4.00 & 2 & 20 \\
pick up the cream cheese and place it in the basket & 20 & 20 & 9.50 & 0.42 & 0.59 & 3.10 & 5 & 20 \\
put the wine bottle on top of the cabinet & 20 & 20 & 9.70 & 0.44 & 0.62 & 3.70 & 3 & 20 \\
pick up the bbq sauce and place it in the basket & 20 & 20 & 9.95 & 0.34 & 0.53 & 4.05 & 6 & 20 \\
pick up the alphabet soup and place it in the basket & 20 & 20 & 10.40 & 0.38 & 0.58 & 4.15 & 4 & 20 \\
push the plate to the front of the stove & 20 & 20 & 10.70 & 0.48 & 0.60 & 3.75 & 3 & 18 \\
pick up the book and place it in the back compartment of the caddy & 20 & 20 & 11.50 & 0.39 & 0.56 & 3.55 & 4 & 20 \\
put both the cream cheese box and the butter in the basket & 20 & 20 & 11.75 & 0.57 & 0.65 & 2.95 & 4 & 20 \\
turn on the stove and put the moka pot on it & 20 & 20 & 12.00 & 0.49 & 0.61 & 4.05 & 2 & 18 \\
open the top drawer and put the bowl inside & 20 & 20 & 12.15 & 0.38 & 0.53 & 5.85 & 2 & 18 \\
put both the alphabet soup and the tomato sauce in the basket & 20 & 20 & 12.50 & 0.61 & 0.70 & 3.05 & 2 & 20 \\
pick up the black bowl on the stove and place it on the plate & 20 & 20 & 13.10 & 0.41 & 0.59 & 4.50 & 3 & 20 \\
put both the alphabet soup and the cream cheese box in the basket & 20 & 20 & 13.20 & 0.60 & 0.70 & 3.15 & 3 & 19 \\
pick up the black bowl next to the ramekin and place it on the plate & 20 & 20 & 13.60 & 0.40 & 0.56 & 4.65 & 2 & 20 \\
pick up the black bowl on the ramekin and place it on the plate & 20 & 20 & 13.70 & 0.41 & 0.59 & 4.90 & 2 & 20 \\
pick up the black bowl on the wooden cabinet and place it on the plate & 20 & 20 & 13.90 & 0.44 & 0.64 & 4.55 & 2 & 20 \\
put the yellow and white mug in the microwave and close it & 20 & 20 & 13.90 & 0.51 & 0.63 & 4.20 & 2 & 19 \\
pick up the black bowl next to the plate and place it on the plate & 20 & 20 & 13.95 & 0.37 & 0.56 & 4.70 & 3 & 20 \\
put the black bowl in the bottom drawer of the cabinet and close it & 20 & 20 & 14.45 & 0.41 & 0.58 & 4.90 & 3 & 19 \\
pick up the black bowl next to the cookie box and place it on the plate & 20 & 20 & 14.65 & 0.45 & 0.60 & 4.30 & 2 & 20 \\
pick up the black bowl on the cookie box and place it on the plate & 20 & 20 & 14.65 & 0.45 & 0.61 & 4.80 & 3 & 19 \\
pick up the black bowl from table center and place it on the plate & 20 & 20 & 14.75 & 0.38 & 0.50 & 5.35 & 2 & 20 \\
pick up the black bowl between the plate and the ramekin and place it on the plate & 20 & 20 & 16.25 & 0.45 & 0.63 & 5.10 & 4 & 20 \\
put the white mug on the plate and put the chocolate pudding to the right of the plate & 20 & 20 & 17.25 & 0.66 & 0.71 & 3.35 & 2 & 19 \\
pick up the black bowl in the top drawer of the wooden cabinet and place it on the plate & 20 & 20 & 17.70 & 0.45 & 0.63 & 5.35 & 3 & 20 \\
put the white mug on the left plate and put the yellow and white mug on the right plate & 20 & 20 & 18.50 & 0.68 & 0.72 & 2.45 & 3 & 19 \\
\midrule
\textbf{All 40 base instructions} & \textbf{800} & \textbf{800} & -- & \textbf{0.44} & \textbf{0.60} & \textbf{3.84} & \textbf{125} & \textbf{787} \\
\bottomrule
\end{tabular}
\caption{\textbf{Per-instruction coverage of the paraphrase set}, ordered by mean paraphrase length. \textbf{Var.}/\textbf{Uniq.}: number of generated and distinct paraphrases; \textbf{Words}: mean paraphrase length in tokens; \textbf{Jacc.}: mean word-level Jaccard similarity to the base instruction; \textbf{Lev.}: mean character-level Levenshtein similarity ratio; \textbf{New}: mean number of tokens not present in the base instruction; \textbf{Reg.}: paraphrases carrying an explicit politeness or register marker; \textbf{Verb}: paraphrases in which the main verb or command surface form differs from the base instruction.}
\label{tab:instrcov}
\end{table*}

\end{document}